%% file: main.tex
\documentclass[journal]{IEEEtran}
\usepackage{bbm}
\usepackage{graphicx}
\usepackage{epsfig,subfigure}
\usepackage{amssymb,amsmath}
\usepackage{enumerate}
\usepackage{latexsym}
\usepackage{booktabs}
\usepackage{multirow} 
\usepackage{array} 
\usepackage{cite}
\usepackage{comment}
\usepackage{todonotes}
\usepackage{authblk}
\usepackage{url}
\usepackage{algorithm}
\usepackage{algorithmic}
\usepackage{caption}
\usepackage{amsmath}
\usepackage{float}
\usepackage{hyperref}

\DeclareMathOperator*{\argmax}{argmax} 

\usepackage{makecell}

\usepackage{graphicx}
\graphicspath{{/Users/ammarhaydari/Documents/Academic_Files/IEEEJournal/Figures/}}

\begin{document}
\title{Deep Reinforcement Learning for Intelligent Transportation Systems: A Survey}
\author{Ammar Haydari, \IEEEmembership{Student Member, IEEE}, 
Yasin Yilmaz, \IEEEmembership{Member, IEEE}
 }

\maketitle

\input{abstract.tex}
\input{Introduction.tex}
\input{Deep_RL.tex}

\input{DRL_on_TLC.tex}
\input{DRL_prop_TLC.tex}
\input{Others.tex}

\bibliographystyle{IEEEtran}
\bibliography{References.bib}

\begin{IEEEbiography}
[{\includegraphics[width=1in,height=1.4in,clip,keepaspectratio]{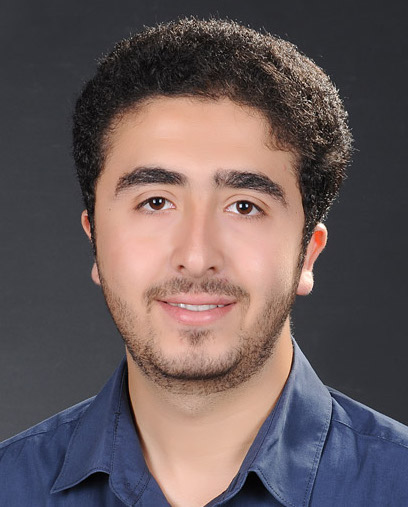}}]{Ammar Haydari}
received the B.Sc. degree in Electrical Engineering from Uludag University, Istanbul, Turkey, in 2014. He is currently a M.Sc. student at the Electrical Engineering Department at the University of South Florida, Tampa. His research interests include intelligent transportation systems and machine learning. 
\end{IEEEbiography}

\begin{IEEEbiography}
[{\includegraphics[width=1in,height=1.4in,clip,keepaspectratio]{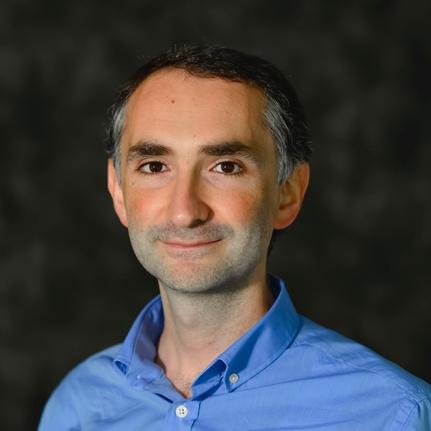}}]{Yasin Y{\i}lmaz}
(S'11-M'14) received the Ph.D. degree in Electrical Engineering from Columbia University, New York, NY, in 2014. He is currently an Assistant Professor of Electrical Engineering at the University of South Florida, Tampa. He received the Collaborative Research Award from Columbia University in 2015. His research interests include statistical signal processing, machine learning, and their applications to cybersecurity, cyber-physical systems, IoT networks, transportation systems, energy systems, and communication systems. 
\end{IEEEbiography}

\end{document}

%% file: abstract.tex
\begin{abstract}
    
Latest technological improvements increased the quality of transportation. New data-driven approaches bring out a new research direction for all control-based systems, e.g., in transportation, robotics, IoT and power systems. Combining data-driven applications with transportation systems plays a key role in recent transportation applications. In this paper, the latest deep reinforcement learning (RL) based traffic control applications are surveyed. Specifically, traffic signal control (TSC) applications based on (deep) RL, which have been studied extensively in the literature, are discussed in detail. Different problem formulations, RL parameters, and simulation environments for TSC are discussed comprehensively. In the literature, there are also several autonomous driving applications studied with deep RL models. Our survey extensively summarizes existing works in this field by categorizing them with respect to application types, control models and studied algorithms. In the end, we discuss the challenges and open questions regarding deep RL-based transportation applications.

\end{abstract}


%% file: Introduction.tex
\section{Introduction}

With increasing urbanization and latest advances in autonomous technologies, transportation studies evolved to more intelligent systems, called intelligent transportation systems (ITS). Artificial intelligence (AI) tries to control systems with minimal human intervention. Combination of ITS and AI provides effective solutions for the 21st century transportation studies. The main goal of ITS is providing safe, effective and reliable transportation systems to participants. For this purpose, optimal traffic signal control (TSC), autonomous vehicle control, traffic flow control are some of the key research areas.  

The future transportation systems are expected to include full autonomy such as autonomous traffic management and autonomous driving. Even now, semi-autonomous vehicles occupy the roads and the level of autonomy is likely to increase in near future. There are several reasons why authorities want autonomy in ITS such as time saving for drivers, energy saving for environment, and safety for all participants. Travel time savings can be provided by coordinated and connected traffic systems that can be controlled more efficiently using self-autonomous systems. When vehicles spend more times on traffic, fuel consumption increases, which has environmental and economic impacts. Another reason why human intervention is tried to be minimized is the unpredictable nature of human behavior. It is expected that autonomous driving will decrease traffic accidents and increase the quality of transportation. For all the reasons stated above, there is a high demand on various aspects of autonomous controls in ITS. One popular approach is to use experience-based learning models, similar to human learning.

Growing population in urban areas causes a high volume of traffic, supported by the fact that the annual congestion cost for a driver in the US was 97 hours and \$1,348 in 2018 \cite{cookson2018inrix}. Hence, controlling traffic lights with adaptive modules is a recent research focus in ITS. Designing an adaptive traffic management system through traffic signals is an effective solution for reducing the traffic congestion. The best approach for optimizing traffic lights is still an open question for researchers, but one promising approach for optimum TSC is to use learning-based AI techniques. 

There are three main machine learning paradigms. Supervised learning makes decision based on the output labels provided in training. Unsupervised learning works based on pattern discovery without having the pre-knowledge of output labels. The third machine learning paradigm is reinforcement learning (RL), which takes sequential actions rooted in Markov Decision Process (MDP) with a rewarding or penalizing criterion. RL combined with deep learning, named deep RL, is currently accepted as the state-of-the art learning framework in control systems. While RL can solve complex control problems, deep learning helps to approximate highly nonlinear functions from complex dataset. 

Recently, many deep RL based solution methods are presented for different ITS applications. There is an increasing interest on RL based control mechanisms in ITS applications such as traffic management systems and autonomous driving applications. Gathering all the data-driven ITS studies related to deep RL and discussing such applications together in a paper is needed for informing ITS researchers on deep RL, as well as deep RL researchers on ITS.

In this paper, we review the deep RL applications proposed for ITS problems, predominantly for TSC. Different RL approaches from the literature are discussed. TSC solutions based on standard RL techniques have already been studied before the invention of deep RL. Hence, we believe standard RL techniques also have high importance in reviewing the deep RL solutions for ITS, in particular TSC. Since traffic intersection models are mainly connected and distributed, multi-agent dynamic control techniques, which are also extensively covered in this survey, play a key role in RL-based ITS applications. 

\subsection{Contributions}

This paper presents a comprehensive survey on deep RL applications for ITS by discussing a theoretical overview of deep RL, different problem formulations for TSC, various deep RL applications for TSC and other ITS topics, and finally challenges with future research directions. The targeted audience are the ITS researchers who want to have a jump start in learning deep RL techniques, and also deep RL researchers who are interested in ITS applications. We also believe that this survey will serve as ``a compact handbook of deep RL in ITS" for more experienced researchers to review the existing methods and open challenges. Our contributions can be summarized as follows.

\begin{itemize}
    \item The first comprehensive survey of RL and deep RL based applications in ITS is presented. 

    \item From a broad concept, theoretical background of RL and deep RL models, especially those which are used in the ITS literature, are explained.

    \item Existing works in TSC that use RL and deep RL are discussed and clearly summarized in tables for appropriate comparisons.

    \item Similarly, different deep RL applications in other ITS areas, such as autonomous driving, are presented and summarized in a table for comparison.
\end{itemize}

\subsection{Organization}

The paper organized as follows.

\begin{itemize}
    \item Section \ref{s:related}: Related Work
    \item Section \ref{s:overview}: Deep RL: An Overview
    \begin{itemize}
        \item Section \ref{s:RL}: Reinforcement Learning
        \item Section \ref{s:deepRL}: Deep Reinforcement Learning
        \item Section \ref{s:summary}: Summary of Deep RL
    \end{itemize}
    \item Section \ref{s:deepRLsettings}: Deep RL Settings for TSC
    \begin{itemize}
        \item Section \ref{s:state}: State
        \item Section \ref{s:action}: Action
        \item Section \ref{s:reward}: Reward 
        \item Section \ref{s:NN}: Neural Network Structure
        \item Section \ref{s:simulation}: Simulation Environments
    \end{itemize}
    \item Section \ref{s:deepRLtsc}: Deep RL Applications for TSC
    \begin{itemize}
        \item Section \ref{s:rlapp}: Standard RL Applications
        \item Section \ref{s:deepRLapp}: Deep RL Applications
    \end{itemize}
    \item Section \ref{s:deepRLothers}: Deep RL for Other ITS Applications
    \begin{itemize}
        \item Section \ref{s:autonomous}: Autonomous Driving
        \item Section \ref{s:energy}: Energy Management
        \item Section \ref{s:road}: Road Control
        \item Section \ref{s:various}: Various ITS Applications
    \end{itemize}
    \item Section \ref{s:challenges}: Challenges and Open Research Questions
\end{itemize}

\section{Related Work}
\label{s:related}

The earliest work summarizing AI models for TSC including RL and other approaches dates back to 2007 \cite{liu2007survey}. At that time, fuzzy logic, artificial neural networks and RL was three main popular AI methods researchers applied on TSC. Due to the connectedness of ITS components, such as intersections, multi-agent models provide a more complete and realistic solution than single-agent models. Hence, formulating the TSC problem as a multi-agent system has a high research potential. The opportunities and research directions of multi-agent RL for TSC is studied in \cite{bazzan2009opportunities}. \cite{mannion2016experimental} discusses the popular RL methods in the literature from an experimental perspective. Another comprehensive TSC survey for RL methods is presented in \cite{yau2017survey}. A recent survey presented in \cite{tong2019artificial} studies AI methods in ITS from a broad perspective. It considers applications of supervised learning, unsupervised learning and RL for vehicle to everything communications. 

\textit{Abduljabbar et al.} summarizes the literature of AI based transportation applications in \cite{abduljabbar2019applications} with three main topics: transportation management applications, public transportation, and autonomous vehicles. In \cite{wei2019survey}, authors discuss the TSC methods in general, including classical control methods, actuated control, green-wave, max-band systems, and RL based control methods. 
\textit{Veres et al.} highlights the trends and challenges of deep learning applications in ITS \cite{veres2019deep}. Deep learning models play a significant role in deep RL. Nonlinear neural networks overcome traditional challenges such as scalability in the data-driven ITS applications. 
Lately, a survey of deep RL applications for autonomous vehicles is presented in \cite{kiran2020deep}, where authors discuss recent works with the challenges of real-world deployment of such RL-based autonomous driving methods. In addition to autonomous driving, in this survey we discuss a broad class of ITS applications where deep RL is gaining popularity, together with a comprehensive overview of the deep RL concept.
  
There is no survey in the literature dedicated to the deep RL applications for ITS, which we believe is a very timely topic in the ITS research. Thus, this paper will fill an important gap for ITS researchers and deep RL researchers interested in ITS.

%% file: Deep_RL.tex
\section{Deep RL: An Overview}
\label{s:overview}

Deep RL is one of the most successful AI models and the closest machine learning paradigm to human learning. It combines deep neural networks and RL for more efficient and stabilized function approximations especially for high-dimensional and infinite-state problems. This section describes the theoretical background of traditional RL and major deep RL algorithms implemented in ITS applications.

\subsection{Reinforcement Learning}
\label{s:RL}

RL is a general learning tool where an agent interacts with the environment to learn how to behave in an environment without having any prior knowledge by learning to maximize a numerically defined reward (or to minimize a penalty). After taking an action, RL agent receives a feedback from the environment at each time step $t$ about the performance of its action. Using this feedback (reward or penalty) it iteratively updates its action policy to reach to an optimum control policy. RL learns from experiences with the environment, exhibiting a trial-and-error kind of learning, similar to human learning \cite{sutton2018reinforcement}. The fundamental trade-off between exploration and exploitation in RL strikes a balance between new actions and learned actions. From a computational perspective, RL is a data-driven approach which iteratively computes an approximate solution to the optimum control policy. Hence, it is also known as approximate dynamic programming \cite{sutton2018reinforcement} which is one type of sequential optimization problem for dynamic programming (DP). 

\begin{figure}[ht]
\centering
\includegraphics[width=.45\textwidth]{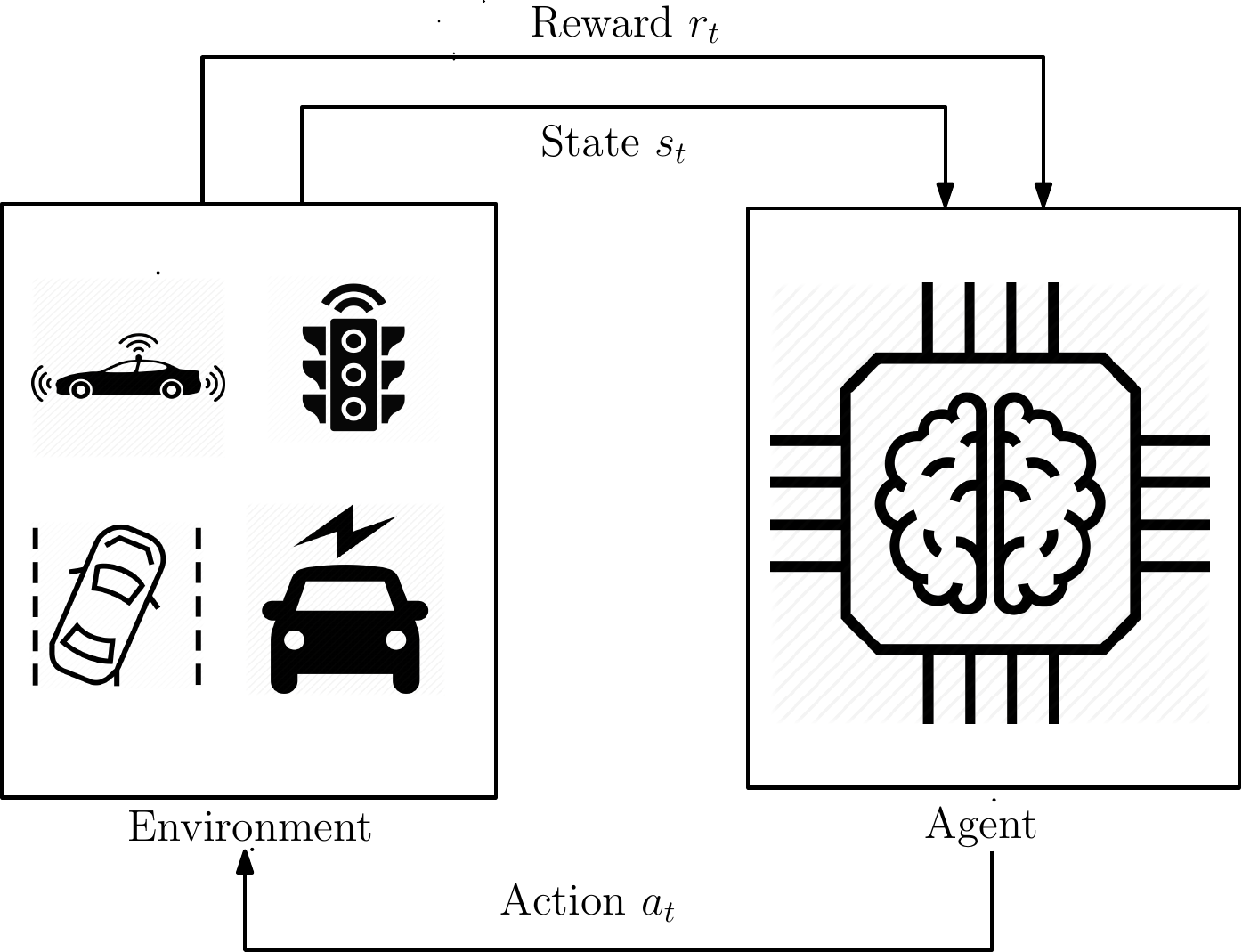}
\caption{Reinforcement learning control loop.}
\label{f:RL_model}
\end{figure}

In a general RL model, an agent controlled with an algorithm, observes the system state $s_t$ at each time step $t$ and receives a reward $r_t$ from its environment/system after taking the action $a_t$. After taking an action based on the current policy $\pi$, the system transitions to the next state $s_{t+1}$. After every interaction, RL agent updates its knowledge about the environment. Fig \ref{f:RL_model} depicts the schematic of the RL process.

\subsubsection{\bf Markov Decision Process}
\label{s:MDP}

RL methodology formally comes from a Markov Decision Process (MDP), which is a general mathematical framework sequential decision making algorithms. MDP consist of 5 elements in a tuple:

\begin{itemize}
    \item A set of states $\mathcal{S}$,
    \item A set of actions $\mathcal{A}$,
    \item Transition function $\mathcal{T}(s_{t+1}|s_t,a_t)$ which maps a state-action pair for each time $t$ to a distribution of next state $s_{t+1}$,
    \item Reward function $\mathcal{R}(s_t,a_t,s_{t+1})$ which gives the reward for taking action $a_t$ from state $s_t$ when transitioning to the next state $s_{t+1}$,
    \item Discount factor $\gamma$ between 0 and 1 for future rewards.
\end{itemize}

The essential Markov property is that given the current state $s_t$, the next state $s_{t+1}$ of system is independent from the previous states $(s_0, s_1,..., s_{t-1})$. In control systems including transportation systems, MDP models are mostly \emph{episodic} in which the system has a terminal point for each episode based on the end time $T$ or the end state $s_T$. The goal of an MDP agent is to find the best policy $\pi^*$ that maximizes the expected cumulative reward $\mathbb{E}[R_t|s,\pi]$ for each state $s$ and cumulative discounted reward (i.e., return)
\begin{equation}
    R_t=\sum_{i=0}^{T-1} \gamma^{i} r_{t+i},
\end{equation}
with the discount parameter $\gamma$ which reflects the importance of future rewards. Choosing a larger $\gamma$ value between 0 and 1 means that agent's actions have higher dependency on future reward. Whereas, a smaller $\gamma$ value results in actions that mostly care about the instantaneous reward $r_t$.

In general, RL agent can act in two ways: (i) by knowing/learning the transition probability $\mathcal{T}$ from state $s_t$ to $s_{t+1}$, which is called model-based RL, (ii) and by exploring the environment without learning a transition model, which is called model-free RL. Model-free RL algorithms are also divided into two main groups as value-based and policy-based methods. While in value-based RL, the agent at each iteration updates a value function that maps each state-action pair to a value, in policy-based methods, policy is updated at each iteration using policy gradient  \cite{sutton2018reinforcement}. We next explain the value-based and policy-based RL methods in detail.

\subsubsection{\bf Value-based RL}
\label{s:valueRL}

Value function determines how good a state is for the agent by estimating the value (i.e., expected return) of being in a given state $s$ under a policy $\pi$ as

\begin{equation}
    {V}^\pi(s)= \mathbb{E}[R_t|s,\pi].
\end{equation}

The optimum value function ${V}^*(s)$ describes the maximized state value function over the policy for all states:
\begin{equation}
    {V}^*(s)= \max_{\pi}{V}^\pi(s), \forall{s} \; \epsilon \; \mathcal{S}.
\end{equation}

Adding the effect of action, state-action value function named as quality function (Q-function) is commonly used to reflect the expected return in a state-action pair:

\begin{equation}
    {Q}^\pi(s,a)= \mathbb{E}[R_t|s,a,\pi].
\end{equation}

Optimum action value function (Q-function) is calculated similarly to the optimum state value function by maximizing its expected return over all states. Relation between the optimum state and action value functions is given by
\begin{equation}
    {V}^*(s)= \max_{a}{Q}^*(s,a), \forall{s} \; \epsilon \; \mathcal{S}.
\end{equation}

Q-function $Q^*(s,a)$ provides the optimum policy $\pi^*$ by selecting the action $a$ that maximizes the $Q$-value for the state $s$:

\begin{equation}
\label{eq:argmax}
    \pi^*(s)= \argmax_{a}{Q}^*(s,a), \forall{s} \; \epsilon \; \mathcal{S}.
\end{equation}

Based on the definitions above, there are two main value-based RL algorithms: Q-learning \cite{watkins1992q} and SARSA \cite{rummery1994line}, which are classified as off-policy RL algorithm, and on-policy RL algorithm, respectively. In both algorithms, the values of state-action pairs ($Q$-value) are stored in a $Q$-table, and are learned via the recursive nature of Bellman equations utilizing the Markov property:

\begin{equation}
    Q^\pi(s_t,a_t)\;=\;\mathbb{E}_{\pi}[r_{t} + \gamma Q^\pi(s_{t+1}, \pi(s_{t+1})]. 
\end{equation}

In practice, $Q^\pi$ estimates are updated with a learning rate $\alpha$ to improve the estimation as follows

\begin{equation}
    Q^\pi(s_t,a_t) \leftarrow Q^\pi(s_t,a_t) + \alpha(y_t - Q^\pi(s_t,a_t))
\end{equation}
where $y_t$ is the temporal difference (TD) target for $Q^\pi(s_t,a_t)$. The TD step size is a user-defined parameter and determines how many experience steps (i.e., actions) to consider in computing $y_t$, the new instantaneous estimate for $Q^\pi(s_t,a_t)$. The rewards $R_t^{(n)}=\sum_{i=0}^{n-1} \gamma^i r_{t+i}$ in the predefined number of $n$ TD steps, together with the Q-value $Q^\pi(s_{t+n},a_{t+n})$ after $n$ steps give $y_t$. The difference between Q-learning and SARSA becomes clear in this stage. Q-learning is an off-policy model, in which actions of the agent are updated by maximizing Q-values over the action, whereas SARSA is an on-policy model, in which actions of the agent are updated according to the policy $\pi$ derived from the $Q$-function:
\begin{equation}
    y_t^{Q-learning} = R_t^{(n)} + \gamma^n \max_{a_{t+n}} Q^\pi(s_{t+n},a_{t+n}),
    \label{eq:q-learning}
\end{equation}
\begin{equation}
    y_t^{SARSA} = R_t^{(n)} + \gamma^n Q^\pi(s_{t+n},a_{t+n}).
\end{equation}
While Q-learning follows a greedy approach to update its Q-value estimates, SARSA follows the same policy for both updating Q-values and taking actions. To encourage exploring new states usually an $\epsilon$-greedy policy is used for taking actions in both Q-learning and SARSA. In the $\epsilon$-greedy policy, a random action is taken with probability $\epsilon$, and the best action with respect to the current policy defined by $Q(s,a)$ is taken with probability $1-\epsilon$. 

In both Q-learning and SARSA, the case with maximum TD steps, typically denoted with $n=\infty$ to express the end of episode, corresponds to a fully experience-based technique called Monte-Carlo RL, in which the Q-values are updated only once at the end of each episode. This means the same policy is used without any updates to take actions throughout an episode. The TD$(\lambda$) technique generalizes TD learning by averaging all TD targets with steps from $1$ to $\infty$ with exponentially decaying weights, where $\lambda$ is the decay rate \cite{sutton2018reinforcement}.


\subsubsection{\bf Policy-based RL}
\label{s:policyRL}

Policy-based RL algorithms treat the policy $\pi_\theta$ as a probability distribution over state-action pairs parameterized by $\theta$. Policy parameters $\theta$ are updated in order to maximize an objective function $J(\theta)$, such as the expected return $\mathbb{E}_{\pi_\theta}[R_t|\theta]=\mathbb{E}_{\pi_\theta}[Q^{\pi_{\theta}}(s_t,a_t)|\theta]$. 
The performance of policy-based methods are typically better than that of value-based methods on continuous control problems with infinite-dimensional action space or high-dimensional problems since policy does not require to explore all the states in a large and continuous space and store them in a table. Although there are some effective gradient-free approaches in the literature for optimizing policies in non-RL methods \cite{rios2013derivative}, gradient-based methods are known to be more useful for policy optimization in all types of RL algorithms. 

Here, we briefly discuss the policy gradient-based RL algorithms, which select actions using the gradient of objective function $J(\theta)$ with respect to $\theta$, called the policy gradient. 
In the well-known policy gradient algorithm REINFORCE \cite{williams1992simple}, the objective function is the expected return, and using the log-derivative trick $\nabla \log \pi_\theta=\frac{\nabla \pi_\theta}{\pi_\theta}$ the policy gradient is written as
\begin{equation}
    \nabla_{\theta}J(\theta) = \mathbb{E}_{\pi_\theta}[Q^{\pi_{\theta}}(s,a) \nabla_\theta \log\pi_\theta].
\end{equation}
Since computing the entire gradient is not efficient, REINFORCE uses the popular stochastic gradient descent technique to approximate the gradient in updating the parameters $\theta$. Using the return $R_t$ at time $t$ as an estimator of $Q^{\pi_{\theta}}(s_t,a_t)$ in each Monte-Carlo iteration it performs the update
\begin{equation}
    \theta \gets \theta + \alpha \nabla_\theta \log\pi_\theta R_t,
\end{equation}
where $\alpha$ is the learning rate. Specifically, $\theta$ is updated in the $\nabla_\theta \log\pi_\theta$ direction with weight $R_t$. That is, if the approximate policy gradient corresponds to a high reward $R_t$, this gradient direction is \emph{reinforced} by the algorithm while updating the parameters. 

One problem with the Monte-Carlo policy gradient is its high variance. To reduce the variance in policy gradient estimates Actor-Critic algorithms use the state value function $V^{\pi_\theta(s)}$ as a baseline. Instead of $Q^{\pi_{\theta}}(s,a)$, the advantage function \cite{baird1994reinforcement} $A^{\pi_\theta}(s,a)=Q^{\pi_\theta}(s,a)-V^{\pi_\theta}(s)$ is used in the policy gradient
\begin{equation}
    \nabla_{\theta}J(\theta) = \mathbb{E}_{\pi_\theta}[A^{\pi_{\theta}}(s,a) \nabla_\theta \log\pi_\theta].
\end{equation}
The advantage function, being positive or negative, determines the update direction: go in the same/opposite direction of actions yielding higher/lower reward than average. Actor-Critic method is further discussed in Section \ref{s:AC} within the Deep RL discussion.

\subsubsection{\bf Multi-Agent RL}
\label{s:multiRL}

Many real world problems require interacting multiple agents to maximize the learning performance. Learning with multiple agents is a challenging task since each agent should consider other agents' actions to reach a globally optimum solution. Increasing the number of agents also increases the state-action dimensions, thus decomposing the tasks between agents is a scalable approach for large control systems. There are two main issues with high-dimensional systems in multi-agent RL in terms of state and actions: stability and adaptation of agents to the environment \cite{busoniu2006multi}. When each agent optimizes its action without considering close agents, the optimal learning for overall system would become non-stationary. There are several approaches to address this problem in multi-agent RL systems such as distributed learning, cooperative learning and competitive learning \cite{busoniu2006multi}.

\subsection{Deep Reinforcement Learning}
\label{s:deepRL}

In high-dimensional state spaces, standard RL algorithms cannot efficiently compute the value functions or policy functions for all states.  
Although some linear function approximation methods are proposed for solving the large state space problem in RL, their capabilities are still up to a certain point. In high-dimensional and complex systems, standard RL approaches cannot learn informative features of the environment for effective function approximation. However, this problem can be easily handled by deep learning based function approximators, in which deep neural networks are trained to learn the optimal policy or value functions. Different neural network structures such as convolutional neural network (CNN) and recurrent neural network (RNN) are used for training RL algorithms in large state spaces \cite{lecun2015deep}. 

The main concept of deep learning is to extract useful patterns from data. Deep learning models are roughly inspired by the multi-layered structure of human neural system. Today, deep learning has applications in a wide spectrum of areas, including computer vision, speech recognition, natural language processing, and the deep RL applications. 

\subsubsection{\bf Deep Q-Network}
\label{s:DQN}

Since value-based RL algorithms learn the Q-function by populating a Q-table, it is not feasible to visit all the states and actions in a large state space and continuous action problems. The leading approach to this problem, called Deep Q-Network (DQN) \cite{mnih2015human}, is to approximate the Q-function with deep neural networks. Original DQN receives raw input image as state, and estimates Q-values from them using CNNs. 
Denoting the neural network parameters with $\theta$ the Q-function approximation is written as $Q(s,a;\theta)$. 
The output of neural network is the best action selected according to (\ref{eq:argmax}) using a discrete set of approximate action values.

The major contribution of \emph{Mnih et al.} \cite{mnih2015human} was two novel techniques to stabilize learning with deep neural networks: target network and experience replay. The original DQN algorithm is shown to significantly outperform the expert human performance on several classic Atari video games. 
The complete DQN algorithm with experience replay and target network is given by Algorithm \ref{alg:dqn}.

\emph{Target Network}: One of the main parts of DQN that stabilize learning is the target network. DQN has two separate networks denoted as the main network that approximates the Q-function, and the target network that gives the TD target for updating the main network. In the training phase, while the main network parameters $\theta$ are updated after every action, target network parameters $\theta^{\textendash}$ are updated after a certain period of time. The reason why target network is not updated after every iteration is that it adjusts the main network updates to keep the value estimations in control. If both networks were updated at the same time, the change in the main network would be exaggerated due to the feedback loop by the target network, resulting in an unstable network. Similar to \eqref{eq:q-learning}, 1-step TD target $y_t$ is written as
\begin{equation}
\label{eq:tdDQN}
    y_t^{DQN} = r_t + \gamma \max_{a_{t+1}} Q^\pi(s_{t+1},a_{t+1};\theta^{\textendash}_t),
\end{equation}
where $Q^\pi(s_{t+1},a_{t+1};\theta^{\textendash}_t)$ denotes the target network.

\emph{Experience Replay}: DQN introduces another distinct feature called experience replay which stores recent experiences $(s_t, a_t, r_t, s_{t+1})$ in replay memory, and samples batches uniformly from the replay memory for training neural network. There are two main reasons why experience replay is used in DQN. Firstly, it prevents the agent from getting stuck into the recent trajectories by doing random sampling since RL agents are prone to temporal correlations in the consecutive samples. Furthermore, instead of learning over full observations, DQN agent learns over mini-batches that increases the efficiency of the training. In a fixed-size memory defined for experience replay, the memory stores only recent $M$ samples by removing the oldest experience for allocating a space to the latest sample. The same technique is applied in other deep RL algorithms \cite{lillicrap2015continuous}, \cite{wang2016sample}.

\emph{Prioritized Experience Replay}: Experience replay technique samples experiences uniformly from the memory, however, some experiences has more impact on learning than the others. A new approach prioritizing significant actions over other actions is proposed in \cite{schaul2015prioritized} by changing the sampling distribution of DQN algorithm. 
The overall idea for prioritized experience replay is that the samples with higher TD error, $y_t^{DQN}-Q^{\pi}(s_t,a_t;\theta_t^-)$, receives higher ranking in terms of probability than the other samples by applying a stochastic sampling with proportional prioritization or rank-based prioritization. The experiences are sampled based on the assigned probabilities.

\begin{algorithm}                      
\caption{DQN algorithm}          
\label{alg:dqn}                           
\begin{algorithmic}[1]                    

\STATE \emph{Input} Replay memory size $M$, batch size $d$, number of episodes $E$, and number of time steps $T$
\STATE {$\emph{Inititalize}$} Main network weights $\theta$
\STATE {$\emph{Inititalize}$} Target network weights $\theta^{-}$
\STATE {$\emph{Inititalize}$} Replay memory 
\FOR{$e=1,\ldots,E$}
    \STATE {$\emph{Inititalize}$} state $s_1$, and action $a_1$
    \FOR{$t=1,\ldots,T$}
    \STATE Take action $a_t = \argmax_a Q^\pi(s_t,a;\theta)$ with probability $1-\epsilon$ or a random action with probability $\epsilon$
    \STATE Get reward $r_t$ and observe next state $s_{t+1}$
    \IF{Replay capacity $M$ is full}
    \STATE Delete the oldest tuple in memory
    \ENDIF
    \STATE Store the tuple $(s_t, a_t, r_t, s_{t+1})$ to replay memory
    \STATE Sample random $d$ tuples from replay memory
    \STATE $y_t=\begin{cases}
    r_t, & \text{if $t=T$}.\\
    r_t + \gamma \max_a Q^\pi(s_{t+1},a_{t+1};\theta^{\textendash}_t), & \text{otherwise}.\end{cases}$
    \STATE Perform policy gradient using $y_t$ for updating $\theta$
    \STATE Update target network every $N$ step, $\theta^{\textendash}=\theta$ 
    \ENDFOR
\ENDFOR
\end{algorithmic}
\end{algorithm}

\subsubsection{\bf Double Dueling DQN} 
\label{s:DDQN}

DQN is the improved version of the standard Q-Learning algorithm with a single estimator. Both DQN and Q-Learning overestimates some actions due to having single $Q$ function estimations. Authors in \cite{van2016deep} proposes doubling the estimators for action selection with main network and action evaluation with target network separately in loss minimization similar to the tabular double Q-learning technique \cite{hasselt2010double}. Instead of selecting the Q value that maximizes future reward using the target network (see Eq. (\ref{eq:tdDQN})), double DQN network selects the action using the main network and evaluates it using the target network. Action selection is decoupled with target network for better Q-value estimation:
 
\begin{equation}
\label{eq:tdDDQN}
    y_t^{DDQN} = r_t + \gamma Q^\pi(s_{t+1},\argmax_{a_{t+1}} Q^\pi(s_{t+1},a_{t+1};\theta);\theta^{\textendash}_t).
\end{equation}

Another improved version of DQN is a dueling network architecture which estimates state value function $V^\pi(s)$ and advantage function $A^\pi(s,a)$ separately for each action \cite{wang2015dueling}. Output of the combination of these two networks is a Q-value for a discrete set of actions through an aggregation layer. This way dueling DQN learns the important state values without their corresponding effects on the actions since state value function $V^\pi(s)$ is an action-free estimation.

These two doubling and dueling models on DQN algorithm with prioritized experience replay are accepted as the state-of-the-art for discrete action-based deep RL.  

\subsubsection{\bf Actor Critic Methods}
\label{s:AC}

\begin{figure}[ht]
\centering
\includegraphics[width=.45\textwidth]{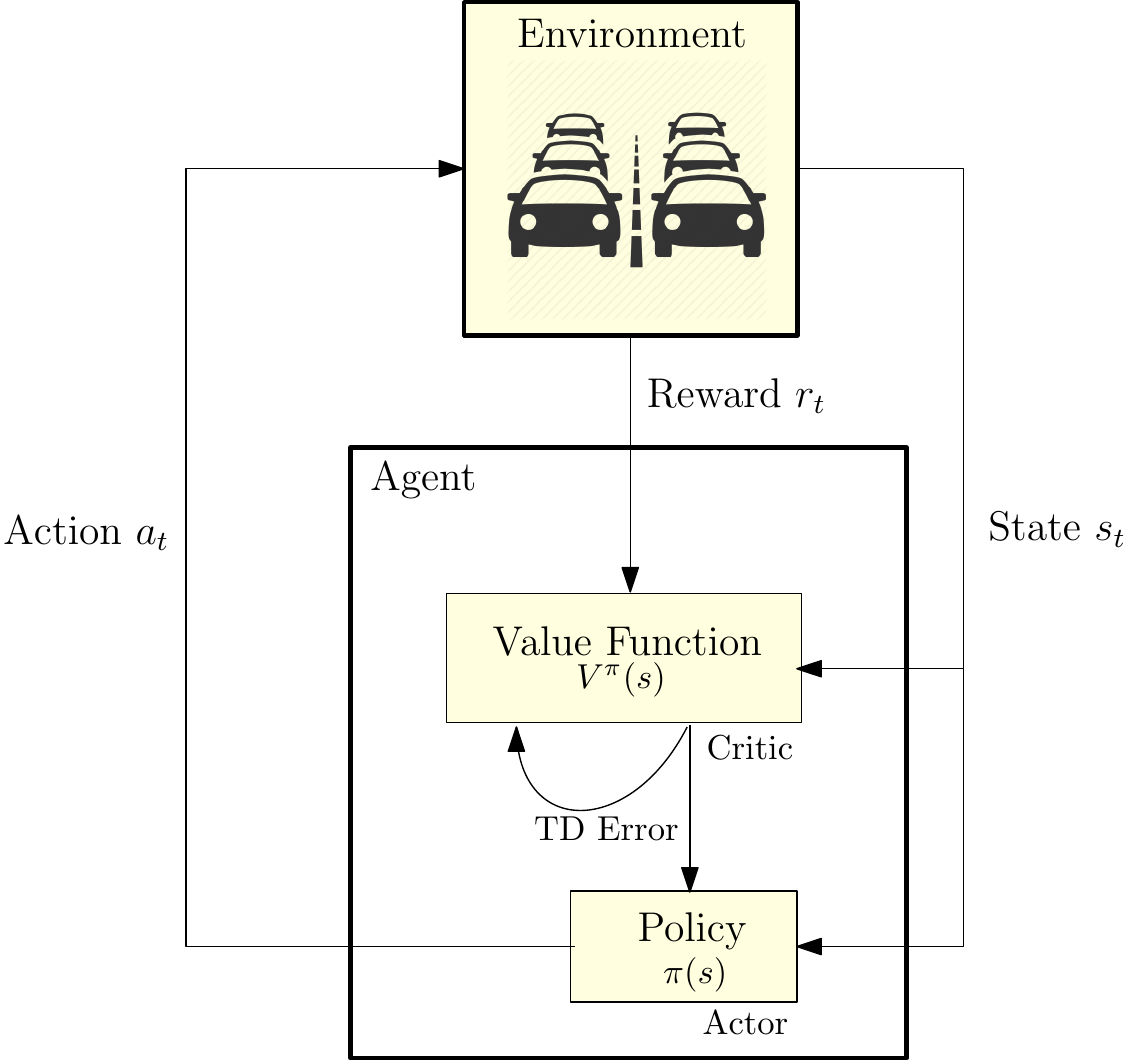}
\caption{Actor Critic control loop.}
\label{f:AC_Model}
\end{figure}

Actor-critic RL models are in between policy-based algorithms and value-based algorithms due to having two estimators: actor using Q-value estimation and critic using state value function estimation (see Fig. \ref{f:AC_Model}).  While actor controls the agent's behaviors based on policy, critic evaluates the taken action based on value function. There are recent papers that deal with the variations of actor-critic models using the deep RL approach \cite{lillicrap2015continuous}, \cite{wang2016sample}, \cite{o2016combining}, in which function approximators for both actor and critic are based on deep neural networks.

Standard DQN techniques with single network estimator are suitable for low-dimensional discrete action spaces. A recent actor-critic algorithm called deep deterministic policy gradient (DDPG) is introduced for solving high-dimensional continuous control problems with deterministic policy gradient approach estimating over state space instead of stochastic policy gradient estimating over state and action spaces together \cite{lillicrap2015continuous}. One of the differences of DDPG from standard DQN is that it uses a new soft target update model doing frequent soft updates.

\subsubsection{\bf Asynchronous Methods}
\label{s:AM}

Improvements in hardware systems allowed RL researchers to perform parallel computing with multiple CPUs or GPUs, which increases the learning pace. First parallel models tested on DQNs advanced the agent performance in terms of lower training time and higher convergence results. For instance, the asynchronous multiple actor-learner model proposed in \cite{mnih2016asynchronous} achieve very high performance in both continuous and discrete action spaces. Multiple actor-learners enable the RL agent to explore the environment with different exploration rates. Furthermore, asynchronous updates do not require replay memory, and learners use accumulated multiple gradients of all experiments done in a predefined update period $T$. Asynchronous advantage actor-critic (A3C), a state-of-the-art deep RL algorithm, updates policy and value networks asynchronously over parallel processors. Each network is separately updated within the update period $T$, and the shared main network is updated with respect to the parameters $\theta^\pi$ and $\theta^V$. The synchronous and simpler version of A3C is known as advantage actor-critic (A2C). 

\subsection{Summary of Deep RL}
\label{s:summary}

In this section, we discussed the background of deep RL, including policy-based and value-based RL models. Before discussing the details of deep RL applications in ITS, it is worth mentioning that certain deep RL algorithms are preferred in different applications depending on the specifications of application domain. While developing new deep RL techniques is an active research area, Q-learning based DQN and actor-critic based DDPG algorithms continue to dominate the RL-based ITS controllers. For high-dimensional state spaces, deep RL methods are preferred over standard RL methods. With regard to action space, policy-based deep RL methods are more suitable for continuous action spaces than value-based deep RL methods. For discrete action spaces, ITS controllers typically use DQN and its variants due to their simpler structures compared to policy-based methods. In general, we can say that Q-learning based DQN models are typically used for less complicated systems which have limited state and action spaces, whereas policy-based or actor-critic algorithms are preferred mainly for large complicated systems, including multi-agent control systems. We should also note here that in many cases the designer can choose between discrete and continuous state and action spaces while setting up the problem. For instance, in TSC, as discussed in the following section, some authors define a continuous action as how much time to extend green light while some other authors define a discrete action space as choosing the green light direction.

%% file: DRL_on_TLC.tex
\section{Deep RL Settings for TSC}
\label{s:deepRLsettings}

Up to this point, we discussed the importance of AI in traffic systems and theoretical background of RL, in particular deep RL. One of the main application areas of deep RL in ITS is controlling signalized intersections. Since most of the existing works are application-oriented, proposed methods differ from each other in various aspects -- e.g., applying deep RL to different intersection models with different technology to monitor traffic, characterizing the RL model with different state-action-reward representations, and using different neural network structures. 
Hence, a direct performance comparison between them is usually not possible.

In these applications, a learning algorithm (deep RL in our case) is implemented in the TSC center to control traffic signals adaptive to the traffic flow. First, the control unit collects the state information, which can be in different formats such as queue length, position of vehicles, speed of vehicles etc., and then control unit takes an action based on the current policy of proposed deep RL method. Finally, agent (control unit) gets a reward with respect to the taken action. By following these steps agent tries to find an optimal policy in order to minimize the congestion on intersection. 

Dealing with the TSC problem on simulators by using RL algorithms requires a good problem formulation in several parts: state, action, reward definitions and neural network structure. In this section, we will discuss these main deep RL configurations together with the traffic simulators used in the literature. 

\subsection{State}
\label{s:state}

The learning performance is highly dependent on an accurate and concrete state definition. Therefore, there are many different state representations used for RL applications on traffic lights. Authors in \cite{mousavi2017traffic} and \cite{garg2018deep} considered raw RGB images as a state representation following the same approach as the original DQN \cite{mnih2015human}. Another similar image-like state representation takes the snapshot of the controlled intersection for forming position and speed of the vehicles \cite{liang2019deep}. Image-like representation format, called discrete traffic state encoding (DTSE), is one of the most popular state definitions in the TSC applications \cite{genders2016using, van2016coordinated, gao2017adaptive, liu2017cooperative, shi2018deep, shabestary2018deep, choe2018deep, garg2018deep, wei2018intellilight, calvo2018heterogeneous}. The reason why researchers prefer to use DTSE is that it acquires the highest available resolution and a realistic set of information from the intersection. Considering $n$ lanes in an intersection, each intersection is divided into cells whose size is on average one vehicle starting from the stopping point of intersection to $m$ meters back. Speed and position of vehicles, signal phases, and accelerations are shown in separate arrays in DTSE. Different variations of those four input types are selected by different researcher. For example, while some researchers select speed and position together \cite{genders2016using, gao2017adaptive}, some others select only one of the four input types for the state representation, such as the position of vehicles \cite{van2016deep, garg2018deep}. 
While DTSE considers the lane characteristics only, \cite{liang2019deep} considers full camera view that also includes road side information in the state definition. Today, many intersections have high quality cameras monitoring the traffic in intersection. To enable DTSE-type state representation, these equipment can be easily extended for monitoring the roads connecting to intersections. 

\begin{figure*}[ht]
\centering
\includegraphics[width=1\textwidth]{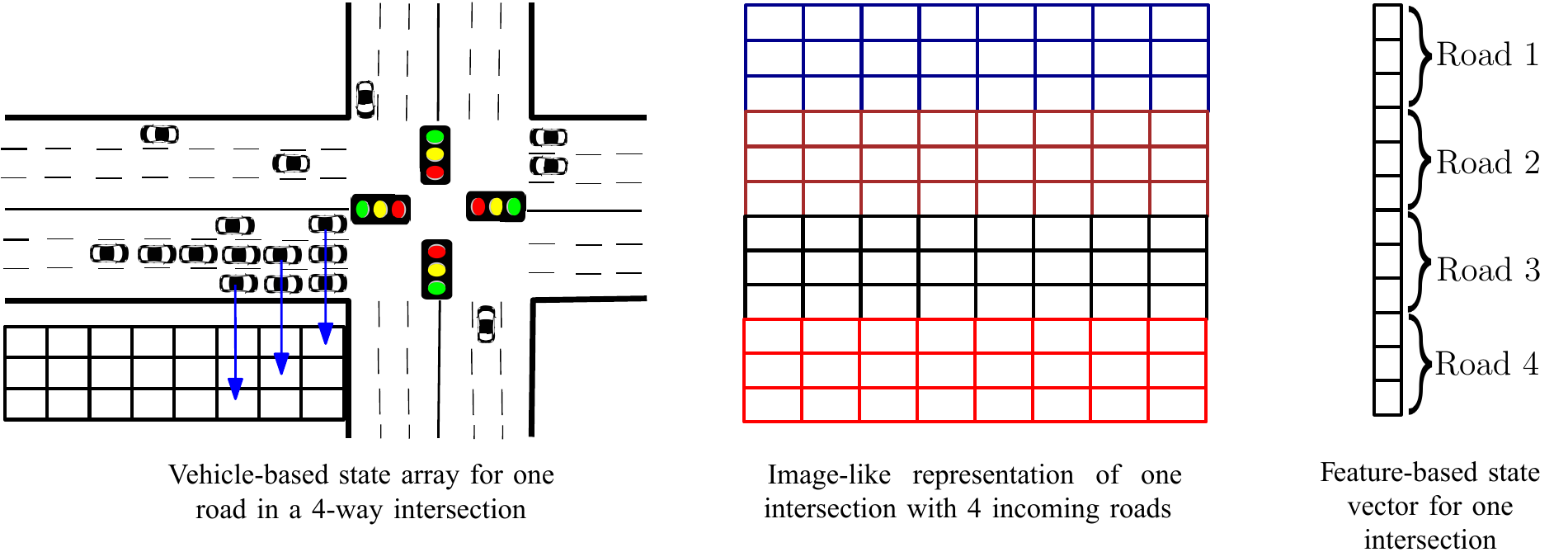}
\caption{Two popular types of state representation: DTSE matrix (middle) and feature-based vector (right). Left figure shows the traffic model with the corresponding vehicle-based state array. In each cell, one vehicle is represented. The matrix in the middle shows a full matrix for one intersection with each road in different colors. Right figure is a feature-based state vector, where each cell represents a lane.}
\label{f:staterep}
\end{figure*}

Another common approach for state representations is forming a feature-based value vector. Instead of vehicle-based state representation, in feature-based state form, average or total value of specific information for each lane is represented on a vector. Queue length, cumulative waiting time in a phase cycle, average speed on a lane, phase duration (green, red, yellow), and number of vehicles in each lane are some of the common features used for state representation. Typically, a combination of such information is collected from intersection \cite{li2016traffic, casas2017deep, lin2018efficient, jang2018agent}. One advantage of such information is that they can be easily collected by road sensors or loop detectors. There are also some other unique traffic features that are not commonly used by researchers such as scoring based on the max speed on lane detectors \cite{casas2017deep}, signal control threshold metrics \cite{zhouerl}, and left turn occupations \cite{wan2018value}. Two common forms of state representations, DTSE and feature vector, are shown in Fig \ref{f:staterep}.

For TSC models with multiple intersections, state definitions also include neighboring traffic light information such as signal phase, number of vehicles, and average speed \cite{liu2017cooperative,zhouerl,ge2019cooperative}.

\subsection{Action}
\label{s:action}

The action taken by the RL algorithm from a set of possible actions after receiving the state has a critical impact on learning. 
In a single four-way intersection, each direction is controlled with green, red and yellow phases. There are several common action selections for a single intersection. The most common one is choosing one of the possible green phases. Another one is the binary action selection keeping the same phase or changing the direction. Third and relatively less common action model is updating the phase duration with a predefined length.  

For a single intersection, mostly there are 4 possible green phases; North-South Green (NSG), East-West Green (EWG), North-South Advance Left Green (NSLG), East-West Advance Left Green (EWLG). During the green phase for a direction, vehicles proceed through the intersection to the allowed direction. When the action selection setting is to select one of the possible green phases, deep RL agent selects an action from these four green phases at each time $t$. After following the yellow and red transitions, the chosen action is performed on traffic lights. Successful agent learning and safety traffic also depend on right red and yellow phase definitions. The early applications simplify the phase definitions to two green phases only, North-South Green (NSG) and East-West Green (EWG) \cite{van2016coordinated, li2016traffic} ignoring the left turns. Another action selection model is binary action, in which green phase interval length is defined beforehand, and at each time $t$, agent decides to either maintain the same phase or proceed to the next phase in a predefined cycle, e.g., NSG $\to$ EWG $\to$ NSLG $\to$ EWLG. When agent selects the action to change the phase, before executing the next green phase, yellow and red transition phases are executed first to have a safe traffic flow \cite{gao2017adaptive, lin2018efficient, choe2018deep, wei2018intellilight, xu2018network}. 

Most of the applications consider discrete action selection from a set of actions, however there are also a few applications that consider continuous outputs \cite{lillicrap2015continuous}, that only controls the duration of the next phase. This type of action definition mostly suitable for multiple intersections. Based on the predefined min and max phase duration, algorithm predicts a time length for the current phase \cite{casas2017deep, genders2018deep}. 

\subsection{Reward}
\label{s:reward}

States in RL can be a feature vector or a high-dimensional matrix, and similarly actions can be a continuous value or a vector of discrete choices. However, reward is always a scalar value which is a function of the traffic data. The role of reward in RL is analyzing the quality of taken action with respect to the state, i.e., penalizing or awarding the agent for the corresponding action. Waiting time, cumulative delay, and queue length are the most common reward definitions in TSC. Waiting time is given by the sum of the times that vehicles are stopped. Delay is the difference between the waiting times of continuous green phases. Queue length is calculated for each lane in an intersection. A special congestion function in transportation planning defined by U.S. Bureau of Public Roads (BPR) is used in some works for the reward definition \cite{liu2017cooperative ,xu2018network}. 
While in some works absolute value of the traffic data is used as a reward, in some others negative value or average value are also used.

\subsection{Neural Network Structure}
\label{s:NN}

The structure of deep neural networks has a high impact on learning in deep RL. Thus, different neural network structures are proposed for TSC in the literature. Multi-layer perceptron (MP), i.e., the standard fully connected neural network model, is a useful tool for classic data classification. An extension of multi-layer perceptron with kernel filters is convolutional neural network (CNN), which provide high performance on mapping image to an output. Standard DQN considers CNN that uses consecutive raw pixel frames for state definition. There are many TSC papers that use CNNs for DTSE state definitions (see Fig. \ref{f:staterep}), e.g., \cite{genders2016using,van2016deepRL,gao2017adaptive}. Residual networks (ResNet) are used to deal with the overfitting problem in CNN-based deep network structures \cite{liu2017cooperative}. Another convolution-based network structure for operations in graphs is graph convolutional networks (GCN).
Recurrent neural networks (RNN), e.g., Long Short-Term Memory (LSTM), are designed to work with sequential data. Since in TSC controlling is done sequentially, RNN is also used in deep RL settings \cite{shi2018deep,choe2018deep}. Another type of neural network model is autoencoder that learns an encoding for high-dimensional input data in a lower-dimensional subspace. The encoded input can be decoded to reconstruct the input, which is commonly used for clearing the noise on input data \cite{li2016traffic}.

\subsection{Simulation environments}
\label{s:simulation}

RL and deep RL applications for TSC are mostly performed on traffic simulators due to life-threatening conditions in real-world experiments. Some authors also use real datasets for experimental study, but still they create a simulation environment based on the real data \cite{natafgi2018smart}. Microscopic individual vehicle-based simulators have been used throughout the years for ITS applications. The earliest available traffic simulator is the Java-based Green Light District (GLD) traffic simulator \cite{wiering2004simulation}, that was initially proposed for an RL-based TSC problem. Many RL papers perform their experiments on the GLD simulator (see Table \ref{t:MRL}), however the most popular open source traffic simulator is Simulation Urban Mobility (SUMO) \cite{behrisch2011sumo}. Open source platforms allow users to modify the simulator for their purposes freely. SUMO enables users to interact with the environment using Python through the traffic control interface (TraCI) library. Different traffic models can be dynamically simulated, including personal vehicles, public vehicles and pedestrians. AIMSUN is a commercial traffic simulator designed and marketed by Transport Simulation Systems (Spain) \cite{casas2010traffic}. Paramics is one of the well-known traffic simulators distributed by Quadstone Paramics (UK) \cite{cameron1996paramics}. VISSIM \cite{fellendorf2010microscopic} is a simulator preferred by researchers due to its interaction with MATLAB, similar to AIMSUN. 

%% file: DRL_prop_TLC.tex
\section{Deep RL Applications for TSC}
\label{s:deepRLtsc}

This section focuses on (deep) RL studies for adaptive TSC. Summary of the works are shown in separate tables for both RL and deep RL models. We can classify learning-based models into two groups in terms of the number of agents: single agent RL which learns the optimal policy with one agent for the entire TSC network, and multi-agent RL which uses multiple agents in the network for acquiring optimal policy. For both standard RL and deep RL-based TSC works, we will discuss the proposed models 
based on 
their characteristic features such as state, action, reward definitions, and neural network structure. 

\subsection{Standard RL applications}
\label{s:rlapp}

\begin{table*}[ht]
\caption{Outline of Single Agent RL approaches for TSC } 
\centering 
\begin{tabular}{l l c c c c c} 
\hline\hline 
Work & RL method & State & Action & Reward & Result comparison  \\ [0.5ex] 
\hline 
Thorpe et al. \cite{thorpe1996tra} & SARSA & \makecell{Vehicle count \\ Fixed vehicle distance \\ Variable vehicle distance} & Binary phase & Fixed penalty (-1) & \makecell{Fixed-time \\ Different states} \\ [1ex] 
Abudlhai et al. \cite{abdulhai2003reinforcement} &Q-learning & Queue length & Binary phase & Total delay & Fixed-time \\ [1ex] 
Camponogara et al. \cite{camponogara2003distributed} & Q-learning & \makecell{Position \& \# vehicles} & Green \& Red phases  & \# waiting vehicles & \makecell{Random policy \\ Longest queue first} \\ [1ex]
Wen et al. \cite{wen2007stochastic}& SARSA & \# vehicles & Binary phase & Coefficients of state & \makecell{Fixed-time \\ Actuated control}\\ [1ex]
El-Tantawy et al. \cite{el2010agent}& Q-learning & \makecell{\# vehicles / Queue length\\ Queue length \\ Cumulative delay} & Green phases  & Change in cum. delay  & Fixed-time \\ [1ex] 
El-Tantawy et al. \cite{el2014design}& \makecell{ Q-learning \\ SARSA \\ TD error} & \makecell{\# vehicles / Queue length\\ Queue length \\ Cumulative delay} & \makecell{Binary phase \\ Green phases} & \makecell{Immediate delay \\ Cumulative delay \\ Queue length \\ \# stops} & \makecell{Fixed-time \\ Actuated control} \\ [1ex] 
Shoufeng et al. \cite{shoufeng2008q}&  Q-learning & Total delay & Time change in green phase & Total delay & Fixed-time\\ [1ex] 
Toubhi et al. \cite{touhbi2017adaptive}&  Q-learning & Max. residual queue & Green phase duration & \makecell{Queue length \\ Cumulative delay \\ Throughput} & Vehicle demand \\ [1ex] 
\hline 
\end{tabular}
\label{t:sRL} 
\end{table*}

\subsubsection{\bf Single agent RL}
\label{s:saRL}
Optimizing intersections with a learning agent receives high attention from researchers since the second half of 1990s. The agent interacts with a simulation environment to learn an optimum control policy for traffic intersection using an RL algorithm. The ultimate goal is mostly controlling a network of coordinated intersections, but the initial step of this research is targeting how to control a single intersection with RL. 
Now we present some RL-based single intersection studies with their distinct features.
  
Traffic signal control with RL-based machine learning is pioneered by the work \cite{thorpe1996tra}, which applies the model-free SARSA algorithm on a single intersection. In this work, \textit{Thorpe and Anderson} considered two scenarios: a four-lane intersection without yellow transition phase, and a $4\times4$ grid style connected intersections where each intersection learns its own Q values separately. After this initial research, several solutions are proposed for single-intersection and multi-intersection traffic networks, where coordinated multi-agent and multi-objective RL dominate the RL research for adaptive TSC. Another SARSA-based TSC method for a single intersection is proposed by \textit{Wen et al.} \cite{wen2007stochastic} with a stochastic control mechanism considering more realistic traffic situations. A specific state space is introduced in this work by partitioning the number of vehicles into sparse discrete values. Authors showed that their proposed model outperforms the fixed-time controller and actuated controller with respect to number of vehicles in queue. 

In \cite{abdulhai2003reinforcement}, authors proposed a model-free Q-learning algorithm for a single intersection with queue length as the state representation and total delay between two action cycles as the reward function. This is the first paper that proposes a simple binary action model that switches the phase direction only. The results of this work are compared with the fixed-time signal controller in different traffic flow patterns in terms of average vehicle delay. A similar Q-learning based RL model is proposed by \textit{Camponogara and Kraus Jr.} \cite{camponogara2003distributed} based on a distributed Q-learning technique on two intersections by assigning separate Q values for each individual agent.
 
\textit{Abdulhai et al.} \cite{el2010agent}, proposed the first RL-based real intersection scenario in Toronto, Canada by using Q-learning with three different state definitions. First state definition is a two-valued function: number of arriving vehicles to the current green direction and number of queued vehicles in the red direction. Other states are defined as queue length and cumulative delay regardless of traffic light. The variable-phase action model in this work selects a green phase among four possible phases defined for a single intersection (NSG, EWG, NSLG, EWLG) instead of a binary action model in a fixed cycle. The same work is extended to a more general concept discussing several on-policy, off-policy RL algorithms on various state, action, reward definitions in an experimental view \cite{el2014design}. Along with the three state representations and the variable-phase action model in \cite{el2010agent}, authors tested their models with also the binary action model in a fixed green-phase cycle, and four reward functions, which are immediate delay, cumulative delay, queue length and the number of stops. 
Different RL algorithms, namely Q-learning, SARSA, and TD error are tested in different state, action, and reward settings on a single intersection. Further, 
two different multi-intersection configurations, 5 intersections in Toronto downtown and a large-scale network of Toronto downtown, are considered for comparison with fixed-time signal control, and actuated signal control models on Paramics simulator. \textit{Toubhi et al.} \cite{touhbi2017adaptive}, assessed three reward definitions, queue length, cumulative delay, and throughput, with Q-learning on a single intersection. The performance of each reward definition is explored on high demand and low demand traffic patterns. There are some other works that also deal with the single intersection control problem using the Q-learning approach \cite{shoufeng2008q, chin2011exploring}. The summary of the presented works are given in Table \ref{t:sRL}.

\subsubsection{\bf Multi-agent RL}
\label{s:maRL}

\begin{table*}[!htbp]
\caption{Overview of Multi-agent RL approaches for TSC } 
\centering 
\begin{tabular}{l c c c c c c c} 
\hline\hline 
Work & RL method & Solution approach& Scenario &Simulator & Result comparison  \\ [0.5ex] 
\hline 
\hline 

Wiering \cite{wiering2000multi} & Model-based RL & \makecell[l]{Waiting time sharing} & 3 by 2 grid & Not specified  & \makecell[l]{Fixed-time controller\\Random controller\\Largest queue first}\\ [1ex] 
\hline 
Steingrover et al. \cite{steingrover2005reinforcement} & Model-based RL & \makecell[l]{Congestion value\\ sharing} &\makecell{12 mixed \\intersections}&GDL&\makecell{TC-1\cite{wiering2000multi}}\\ [1ex] 
\hline 
I{\v{s}}a et al. \cite{ivsa2006reinforcement} & Model-based RL & \makecell[l]{Congestion \& accident \\value sharing} &\makecell{12 mixed \\ intersections}&GDL&\makecell{TC-1 \cite{wiering2000multi} \\ TC-most \\ Accident car removing}\\ [1ex]
\hline 
Kuyer et al. \cite{kuyer2008multiagent} &Model-based RL& \makecell[l]{Coordination graph based \\max plus} & \makecell{3 intersections \\ 4 intersections \\ 15 mixed intersections} &GDL& \makecell{TC-1 \cite{wiering2000multi} \\ TC-SBC \cite{steingrover2005reinforcement} \\ Max-plus}\\ [1ex]
\hline 
Bakker et al. \cite{bakker2010traffic} &Model-based RL& \makecell[l]{Partially observed MDP} & \makecell{15 mixed intersections }& GDL & \makecell{TC-1 \cite{wiering2000multi}\\Diff. partial \\observation techniques}\\ [1ex] 
\hline 
Houli et al. \cite{houli2010Multiobjective} &Model-based RL & Multi-objective learning&\makecell[l]{Real road map \\ in Beijing}& Paramics& \makecell{Fixed controller\\Actuated control\\Single agent RL}\\ [1ex] 
\hline 
Brys et al. \cite{brys2014distributed} & SARSA &\makecell[l]{Multi-objective learning \\ Tile coding} & 2 by 2 grid& AIM& \makecell{Actuated control \\ Distributed learning \cite{taylor2011distributed}} \\ [1ex] 
\hline 
Khamis et al. \cite{khamis2012multi} &\makecell{Model-based RL \\with Bayesian trans. func.}&Multi-objective learning & \makecell{12 mixed \\intersections} & GLD & TC-1 \cite{wiering2000multi}\\ [1ex] 
\hline 
Khamis et al. \cite{khamis2012enhanced} &\makecell{Model-based RL \\with Bayesian trans. func.}&\makecell{Multi-objective learning\\Agent Cooperation}&\makecell{12 mixed \\intersections} & GLD & TC-1 \cite{wiering2000multi}\\ [1ex] 
\hline 
Khamis et al. \cite{khamis2014adaptive} &\makecell{Model-based RL \\with Bayesian trans. func.\\ Hybrid exploration}&\makecell{Multi-objective learning\\Agent Cooperation}&\makecell{22 mixed \\intersections} & GLD &\makecell{ TC-1 \cite{wiering2000multi} \\ SOTL \cite{cools2013self}}\\ [1ex]
\hline 
Jin et al. \cite{jin2019multi} &\makecell{SARSA with \\function approximators}&\makecell{Multi-objective learning\\Threshold lexicographic\\ ordering}&\makecell{3 intersections \\in Stockholm} & SUMO &\makecell{ Comparison between\\multiple function\\approximators}\\ [1ex]
\hline 
Prashanth et al. \cite{prashanth2011reinforcement} &\makecell{Q-learning \\Actor-critic}&Function approximation& \makecell{2 by 2 grid\\5 intersectins}  &GLD& \makecell{Fixed-time control\\No function approx.}\\ [1ex] 
\hline 
Prashanth et al. \cite{prashanth2011rl}  &\makecell{Q-learning}&Function approximation& \makecell{2 by 2 grid\\ 3 by 3 grid \\5 intersections\\ 9 intersections}  &GLD& \makecell{Fixed-time control\\No function approx.\\SOTL\cite{cools2013self}}\\ [1ex] 
\hline 
Pham et al. \cite{pham2013learning} & SARSA &\makecell{ Tile coding} & 2 by 2 grid& AIM& \makecell{Random RL  \\ Distributed learning \cite{taylor2011distributed}}\\ [1ex] 
\hline 
Abdoos et al. \cite{abdoos2014hierarchical} &Q-learning&\makecell{2-level hierarchical\\ control}&3 by 3 grid&AIMSUN& \makecell{1-level Q-learning}\\ [1ex] 
\hline 
Arel et al. \cite{arel2010reinforcement} &Q-learning& \makecell{Neural networks\\ Hierarchical control}&5 intersections& MATLAB &Longest queue first\\ [1ex] 
\hline 
El-Tantawy et al. \cite{el2012multi} &Q-learning & \makecell{Indirect coordination\\Direct coordination} &5 intersections& Paramics &\makecell{Comp. between\\proposed models} \\ [1ex] 
\hline 
El-Tantawy et al. \cite{el2013multiagent} &Q-learning & \makecell{Indirect coordination\\Direct coordination} &\makecell{Real road map\\in downtown Toronto}& Paramics &\makecell{Fixed-time control\\ Semi-actuated control\\ Full actuated control} \\ [1ex] 
\hline 
Salkham et al. \cite{salkham2008collaborative} &Q-learning& \makecell{Adaptive round robin\\based collaboration} &\makecell{Real road map in \\Dublin City}& UTC &\makecell{Independent RL\\Fixed-time\\ SAT-like\cite{richter2006learning}}\\ [1ex] 
\hline 
Aziz et al. \cite{aziz2018learning} &Av. expected reward & Multi-reward structure& \makecell{8 intersections\\11 intersections} &VISSIM& \makecell{Q-learning\\SARSA\\Fixed-time control\\Adaptive control} \\ [1ex] 
\hline 
Aslani et al. \cite{aslani2017adaptive} & Actor-critic & \makecell{Tile coding \\Radial basis functions}&\makecell{Real road map\\in downtown Tehran}&AIMSUN&\makecell{Q-learning\\Fixed-time control\\Actuated control}\\ [1ex] 
\hline 
Xu et al. \cite{xu2013study} &Q-learning&\makecell{Non-zero sum based\\Markov game}&3 by 3 grid&MATLAB&\makecell{Ind. Q-learning\\Fixed-time control\\Longest queue first}\\ [1ex] 
\hline 
Abdoos et al. \cite{abdoos2011traffic} &Q-learning& State discretization & 50 intersections & AIMSUN & Fixed-time control\\ [1ex] 
\hline 
Balaji et al. \cite{balaji2010urban} &Q-learning&Neighboor cooperation&\makecell{Real road map \\in Singapore}&Paramics& \makecell{Hierarchical MS \cite{keong1993glide}\\Cooperative ensemble\\Actuated control}\\ [1ex] 
\hline 
Cahill et al. \cite{cahill2010soilse} &Q-learning&\makecell{CUSUM-based \\pattern change detection}&\makecell{Real road map in \\Dublin City}&UTC& SAT-like\cite{richter2006learning}\\ [1ex] 
\hline
Araghi et al. \cite{araghi2015distributed}&Q-learning&Distributed learning&3 by 3 grid&Paramics&Fixed-time control\\[1ex] 
\hline 
\end{tabular}
\label{t:MRL} 
\end{table*}

Applying single agent RL algorithms individually on different intersections can be a good solution up to a certain point, however large intersection networks suffer from this approach. A cooperative learning approach is needed to reach an optimum policy over all network. Several multi-agent learning models are proposed for controlling multiple intersections cooperatively.

While most of the RL applications for single intersection consider model-free algorithms, such as Q-learning and SARSA, early multi-agent papers proposed model-based RL algorithms that form a transition probability distribution. A prominent multi-agent RL work for large traffic networks is \cite{wiering2000multi} by \textit{Wiering}, in which three algorithms were proposed, namely TC-1, TC-2, TC-3, based on the coordination between vehicles and intersections considering local and global information for the state function. States are formed based on the traffic light configuration of intersections, position of the vehicles and the destination of the vehicles at each intersection. The approach for creating a state representation in this early work is not realistic due to unknown destination for each vehicle. The proposed models iteratively update value functions to minimize the waiting times of vehicles. 
The results are compared with four standard TSC models: fixed-time control, random control, longest queue first, and most-car model. 
Several works extended the \textit{Wiering}'s approach in different perspectives. For example, \textit{Steingrover et al.} proposed an extension to the TC-1 method by including congestion information on other intersections \cite{steingrover2005reinforcement}. Two different extensions, called TC-SBC and TC-GAC, are proposed by the authors. The former increases the state size by adding congestion values to the state space, whereas the latter uses a congestion factor while computing the value function instead of increasing the state space. \textit{Isa et al.} \cite{ivsa2006reinforcement} proposed a further improvement to the TC-1 method by including congestion and accident information in the state representation, 
which further increases the state representation. 
While the works presented until now does not consider coordination between agents for joint action selection, \textit{Kuyer et al.} introduces a new approach that enables coordination between agents by using the max-plus algorithm \cite{kuyer2008multiagent}. In this model, agents coordinate with each other to reach optimum joint actions in finite iterations. Another multi-agent RL model is proposed by \textit{ Bakker et al.} \cite{bakker2010traffic} with partial observations for the state spaces of connected intersections. 
This case is of interest when the system cannot access the full state information due to some reasons such as faulty sensors. All these works \cite{steingrover2005reinforcement, ivsa2006reinforcement, kuyer2008multiagent, bakker2010traffic} use \textit{Wiering}'s approach \cite{wiering2000multi} as a benchmark.

Multi-objectivity is gaining popularity in RL \cite{liu2015Multiobjective} due to its capabilities in complex environments. When a single objective is selected for the overall traffic system, such as the \textit{Wiering}'s objective which aims to decrease the waiting time of all vehicles, 
it may not serve well the needs of different traffic conditions. Authors in \cite{houli2010Multiobjective} consider a multi-objective approach in their multi-agent RL work for TSC. In particular, vehicle stops, average waiting time, and maximum queue length are targeted as objectives for low, medium, and high traffic volume, respectively. 
Different Q functions are updated with appropriate reward functions in these three traffic conditions. 
\textit{Taylor et al.} proposes a non-RL based basic learning algorithm called Distributed Coordination of Exploration and Exploitation (DCEE) \cite{taylor2011distributed} to tackle the TSC problem. 
Authors in \cite{brys2014distributed} and \cite{pham2013learning} consider a multi-agent RL-based SARSA algorithm with tile coding and compare it with DCEE under different traffic conditions. 

\textit{Khamis et al.} studied multi-objective RL control for traffic signals in three papers \cite{ khamis2012multi,khamis2012enhanced, khamis2014adaptive}. In the first paper \cite{khamis2012multi}, authors considered Bayesian transition probability for model-based RL using several objectives for forming the reward function. The second paper followed the same approach \cite{khamis2012enhanced} with more specific objectives. The third paper 
\cite{khamis2014adaptive} 
extends the previous works to a total of seven objectives with a novel cooperative exploration function and experiments in several road conditions and vehicle demands. The paper also improves the practicality of GLD traffic simulator from different perspectives, e.g., continuous control, probabilistic travel demand. The results of these three papers are compared with TC-1 proposed by \textit{Wiering} \cite{wiering2000multi} and the adaptive SOTL method \cite{cools2013self}. The latest and most compact RL-based multi-objective multi-agent TSC study is presented in \cite{jin2019multi}. In this work, travel delay and fuel consumption are defined as learning objectives for the RL agents, and a specific technique called threshold lexicographic ordering is used for online multi-objective adaptation. SARSA is experimented in this work with several  function approximators, one of which is based on neural networks. It is worth noting that SARSA with Q-value estimation is not considered a deep RL approach since it does not include experience replay and target network tricks discussed in \ref{s:DQN}.

Before DQN was introduced, function approximators were popular for Q-functions with large state space. For instance, authors in \cite{prashanth2011reinforcement, prashanth2011rl} proposed two RL models for TSC using function approximation-based Q-learning and actor-critic policy iteration. The proposed Q-learning method outperforms standard Q-learning with full state representation. A novel neural network-based multi-agent RL for TSC is proposed in \cite{arel2010reinforcement}, which uses local agents and global agents. While local agent controls the traffic lights via the longest queue first algorithm, global agent controls the traffic lights with a neural network-based Q-learning approach, which is very similar to DQN discussed in \ref{s:deepRL}.

Actor-critic-based multi-agent RL is an emerging field that uses continuous state representation. Discretizing the state space is prone to missing information about the state. \textit{Aslani et al.} proposed a continuous space actor-critic control model for multiple intersections \cite{aslani2017adaptive}, in which tile coding and radial basis-based function approximators are presented. Although state space is continuous, action space to determine the duration of the next green phase is discrete. 
In experiments, discrete and continuous state space based actor-critic models are tested in the city of Tehran. In another work, two-layer hierarchical multi-agent RL method is studied \cite{abdoos2014hierarchical}, which implements a single agent for each intersection using Q-learning, and controls a wide area network with function approximator based on tile coding on second layer.

There are several studies offering coordination between the neighbor agents for reaching a joint optimum performance. To this end, \textit{Tantawy et al.} proposed a Q-learning based multi-agent RL approach for road network coordination \cite{el2012multi, el2013multiagent}. RL agents learn the coordination directly or indirectly, called MARLIN-DC and MARLIN-IC. While a small-scale road network is presented in \cite{el2012multi}, in the extended paper \cite{el2013multiagent} authors investigate a large network of 59 intersections in downtown Toronto. \cite{balaji2010urban} presents another coordination-based TSC model implementing distributed Q-learning agents for a large network where neighbour agents share congestion values with each other. Experiments are performed on the Paramics simulation environment using a real traffic network in Singapore with different travel demand configurations. \textit{Xu et al.} \cite{xu2013study} proposed a coordination module based on nonzero-sum Markov game for the multi-agent RL environment. Q-learning is used on each intersection as a single agent, and their coordination is controlled with a Markov game-based mathematical model.

A new technique for multiple intersection environments is proposed in \cite{aziz2018learning} using the R-Markov Average Reward technique and a multi-objective reward definition for RL. The result of this work is compared with fixed-time controller, actuated controller, Q-learning and SARSA on the Paramics simulation environment by simulating an 18-intersection network. \textit{Chu et al.} \cite{chu2016large}, proposed a regional to central multi-agent RL model for large-scale traffic networks. In low traffic density, authors claim that for large-scale networks collaboration is not needed between regions, i.e., learning the traffic model in local region is enough to attain a globally appropriate learning. \textit{Araghi et al.}, \cite{araghi2015distributed} presents a distributed Q-learning based multi-agent RL controller that predicts the green phase duration on the next phase cycle. 
Other multi-agent RL applications are studied in \cite{abdoos2011traffic, cahill2010soilse, salkham2008collaborative}. Table \ref{t:MRL} gives an overview of the multi-agent RL works.

\subsection{Deep RL applications}
\label{s:deepRLapp}
Here we discuss deep RL-based TSC applications considering. A summary of the discussed works is provided in Table \ref{t:deepRL} considering the used deep RL algorithms, network structures, simulation environments, and comparison with benchmarks.

\subsubsection{\bf Single agent deep RL}
\label{s:sadeepRL}
 
 \begin{table*}[!htbp]
\caption{Outline of Deep RL approaches for TSC. } 
\centering 
\begin{tabular}{l c c c c c c } 
\hline\hline 
Work & \makecell{Deep RL \\neural network structure} & Multi-agent &State - DTSE& Scenario &Simulator& Result comparison \\ [1ex] 
\hline\hline 
Genders et al. \cite{genders2016using}& DQN - CNN &No& Yes&\makecell{Single int.}&SUMO & MP(64) DQN\\ [1ex] 
\hline
Van der Pool et al. \cite{van2016deepRL}&DQN - CNN&\makecell{Max-plus \\Transfer planning}&Yes& \makecell{Single int.\\2 intersections\\3 intersections\\ 2 by 2 grid}&SUMO& Model based RL\\ [1ex] 
\hline
Van der Pool et al. \cite{van2016coordinated}&DQN - CNN&\makecell{Max-plus \\Transfer planning}&Yes& \makecell{2 intersections\\3 intersections\\ 2 by 2 grid}&SUMO& Model based RL\\ [1ex]
\hline
Li et al.\cite{li2016traffic}&\makecell{DQN - Autoencoder}&No&No&Single int.&Paramics&Q-learning \\ [1ex]
\hline
Gao et al. \cite{gao2017adaptive}&DQN - CNN &No& Yes&Single int.&SUMO&\makecell{Fixed-time control\\Longest queue first}\\ [1ex]
\hline
Liu et al. \cite{liu2017cooperative}&DQN-ResNeT\cite{he2016deep}&Policy sharing&Yes&2 by 2 grid&SUMO&\makecell{SOTL\\ DQN without CNN\\ Q-learning}\\ [1ex]
\hline
Casas \cite{casas2017deep}&DDPG&\makecell{Multiple actor-\\critic learner} &No&\makecell{Single int.,\\6 intersections,\\Real map from \\Barcelona}&Aumsim&\makecell{Q-learning\\Random}\\ [1ex]
\hline
Shi et al. \cite{shi2018deep}&DQN-RNN&\makecell{Max-plus\\Transfer planning}&Yes&2 by 2 grid&USTCMTS2.1&\makecell{Fixed-time control\\Q-learning}\\ [1ex]
\hline
Mousavi et al. \cite{mousavi2017traffic}&\makecell{DQN-CNN\\ A2C-CNN}&No&Real image&Single int.&SUMO&\makecell{Fixed-time control}\\ [1ex]
\hline
Lin et al. \cite{lin2018efficient}&A2C-CNN&\makecell{Multiple actor-\\critic learners}&No&3 by 3 grid&SUMO&\makecell{Fixed-time control\\Actuated control}\\ [1ex]
\hline
Genders et al. \cite{genders2018evaluating}&A3C-MP& No &Yes&Single int.&SUMO&Actuated control\\ [1ex]
\hline
Shabestary et al. \cite{shabestary2018deep}&DQN-CNN&No&Yes&Single int.&Paramics&\makecell{Different rewards\\Q-learning} \\[1ex]
\hline
Choe et al. \cite{choe2018deep}&DQN-RNN&No&Yes&Single int.&SUMO&CNN-DQN\\ [1ex]
\hline
Garg et al. \cite{garg2018deep}&\makecell{DQN-CNN\\(Policy based)}
&No&Yes&Single int.&Unity3d&\makecell{Fixed-time control\\No traffic light}\\ [1ex]
\hline
Coskun et al. \cite{cocskun2018deep}&\makecell{DQN-CNN\\Actor-Critic}&Joint learning&No&4 intersections&SUMO&DQN(Policy based)\\ [1ex]
\hline
Wei et al. \cite{wei2018intellilight}&DQN-CNN&No&Yes&Single int.&\makecell{SUMO\\Real dataset}&\makecell{Fixed-time\\SOTL}\\ [1ex]
\hline
Natafgi et al. \cite{natafgi2018smart}&DQN-CNN&No&No&Single int.& SUMO&Fixed tim\\ [1ex]
\hline
Nishi et al. \cite{nishi2018traffic}&NFQI-Graph CNN \cite{riedmiller2005neural} &No&No&6 intersections&SUMO&\makecell{Fixed-time\\CNN-DQN}\\ [1ex]
\hline
Wan et al. \cite{wan2018value}&Modified DQN&No&No&Single int.&VISSIM&\makecell{DQN\\Fixed-time control}\\ [1ex]
\hline
Calvo et al. \cite{calvo2018heterogeneous}&DQN-CNN&\makecell{Independent DQN\\fingerprints}&Yes&3 intersections&SUMO&Fixed-time control\\ [1ex]
\hline
Genders \cite{genders2018deep}&DDPG&Multiple learners&No&\makecell{Real map from\\ Luxemburg}&SUMO&Fixed-time control\\ [1ex]
\hline
Chu et al. \cite{chu2019multi}&A2C-RNN&Policy sharing&No&\makecell{5 by 5 grid\\Monaco city map}&SUMO&\makecell{Ind-Q-learning\\Ind-DQN\\Ind-A2C}\\ [1ex]
\hline
Liang et al. \cite{liang2019deep}&\makecell{Double Dueling\\DQN-CNN}&No&Yes&Single int.&SUMO&\makecell{Fixed-time control\\Actuated control\\DQN}\\ [1ex]
\hline
Genders et al. \cite{genders2019asynchronous}&\makecell{Asynchronous n-step\\Q-learning}&No&No&Single int.&SUMO&\makecell{Linear learning\\Actuated control\\Random control}\\ [1ex]
\hline
Zhou et al. \cite{zhouerl}&DQN-MP&Threshold based& No&\makecell{Real map from\\New york city}&SUMO&Diff veh. demands\\ [1ex]
\hline
Xu et al. \cite{xu2018network}&DQN-RNN&\makecell{Critical node\\ discovery}&No&\makecell{20 intersections\\50 intersections\\100 intersections}&SUMO&\makecell{Fixed-time\\ SOTL\\Q-learning\\DQN}\\ [1ex]
\hline
Tan et al. \cite{tan2019cooperative}& \makecell{DQN (Value based)\\DDPG (Wolpertinger)}&\makecell{Hierarchical\\cooperation}&No&\makecell{6 intersections\\12 intersections\\24 intersections}&SUMO&\makecell{Fixed-time control\\Q-learning\\DQN}\\ [1ex]
\hline
Ge et al. \cite{ge2019cooperative}&DQN-CNN&Q value transfer&Yes&\makecell{Heterogeneous 4 int.\\ 2 by 3 grid}&SUMO&\makecell{Dist. Q-learning\\DQN}\\ [1ex]
\hline
Liu et al. \cite{liu2018deep}&DQN-MP&No&No&\makecell{Single int.\\ 4 intersections}&Python&No comparison\\ [1ex]
\hline
Zhang et al. \cite{zhang2018intelligent}&DQN&No&No&\makecell{Arterial topology\\4 by 4 grid}&SUMO& \makecell{Partially Observable\\ states}\\ [1ex]
\hline 
\end{tabular}
\label{t:deepRL} 
\end{table*}

In recent years, deep RL based learning tools for adaptive intersection controls gained a great attention from transportation researchers. 
After researchers proposed several architectures for different traffic scenarios using standard RL in the last two decades, 
invention of deep RL made a huge impact on the ITS research, 
in particular TSC. Due to 
its capability of dealing with large state space, 
a number of deep RL models have been proposed for controlling traffic lights. The deep RL paradigm is basically based on approximating Q-functions with deep neural networks. The earliest work using this approach is \cite{arel2010reinforcement}. Although a neural network-based RL model is proposed in this paper, it is not a full DQN algorithm due to lack of experience replay and target network, which are essential components of DQN 
\cite{mnih2015human}. 

The initial work on controlling traffic signals with a deep RL approach is \cite{genders2016using} by \textit{Genders et al.}. In this work, authors use discrete traffic state encoding model, called DTSE, to form an image-like state representation based on detailed information from the traffic environment. The proposed state model is an input to CNN for approximating the Q-values of discrete actions. The experiments are performed on the SUMO simulation environment with a single intersection where 4 green phases are selected as actions. In order to show the power of CNN on the DTSE state form, the results are compared with Q-learning using a single layer neural network. In \cite{genders2018evaluating}, the same authors studied the effects of different state representations for intersection optimization using the A3C algorithm. Three separate state definitions are experimented on a single intersection using a dynamic traffic environment. The first form of state definition considered in the paper is given by the occupancy and average speed of each lane. The second state definition is the queue length and vehicle density for each lane. The third state form is the image like representation, DTSE, with Boolean position information in which the existence of vehicle is represented with 1. 
The results show that the resolution of state representation does not effect the performance of RL agent in terms of delay and queue length. The same authors, in a recent paper \cite{genders2019asynchronous}, studied asynchronous deep RL model for TSC. In asynchronous n-step Q-learning \cite{mnih2016asynchronous}, the main job is divided to multiple processors, and each processor learns its local optimal parameters individually. Global parameters for the general network is updated after every n-step. The proposed architecture in \cite{genders2019asynchronous} improves the performance almost 40\% compared to the fixed-time and actuated traffic controllers. 

Authors in \cite{li2016traffic}, proposed an autoencoder-based deep RL algorithm for a single intersection with dynamic traffic flow. Autoencoders are considered for action selection by mapping input queue length to a low-dimensional action set. Bottleneck layer, which is the output of decoding part, is used for Q-function approximation. The results are compared with standard Q-learning using the Paramics simulator. Currently, this is the only work in the literature that uses autoencoders to approximate action values. 
In \cite{gao2017adaptive}, \textit{Gao et al.} proposed 
a new neural network architecture in which state is a combination of the speed and position of vehicles based on DTSE. 
The output of neural network is binary action whether to keep the same action or change the action in a predefined phase cycle. The proposed model is compared with the fixed-time controller and the longest queue first controller.  

The authors in \cite{mousavi2017traffic}, presented two deep RL algorithms for controlling isolated intersections: value-based DQN, and policy-based actor critic. The state for both agents is raw consecutive image frames following exactly the same approach with original DQN. As stated in the original paper \cite{mnih2015human}, DQN algorithm suffers from instability issues. \cite{mousavi2017traffic} shows that the policy-based deep RL technique solves this issue by having a smooth convergence and a stable trend after convergence. \textit{Shabestary et al.} \cite{shabestary2018deep} proposed a DQN-based solution for adaptive traffic signal control on an isolated intersection using a new reward definition. The reward and action defined in this paper are change in the cumulative delay and 8 different green phases, as opposed to the commonly used binary action set or 4 green phases for a single intersection. 

\textit{Choe et al.} proposed a RNN-based DQN model in a single intersection TSC scenario \cite{choe2018deep}. It is shown that the performance of RNN-based DQN decreased the travel time compared to the popular CNN structure. A policy gradient-based deep RL method is proposed for adaptive traffic intersection control in \cite{garg2018deep}, which presents experiments on a novel realistic traffic environment called Unity3D by using raw pixels as an input state to policy-based DQN. 
The proposed model has similar results with the fixed-time intersection control model. An action value-based DQN with a novel discount factor is proposed by \textit{C. Wan et al.} \cite{wan2018value}. The proposed dynamic discount factor takes execution time into account with the help of infinite geometric series. The proposed model is tested on a single intersection using the SUMO simulator by comparing it with the fixed-time controller and the standard DQN-based controller. 

A new DQN-based controller, called IntelliLight, with a new network architecture is described in \cite{wei2018intellilight}. The reward function consist of multiple components: sum of the queue length over all lanes, sum of delay, sum of waiting time, traffic light state indicator, number of vehicles that passed the intersection since the last action, and sum of travel times since the last action. The proposed method is experimented on SUMO using a single intersection. A real dataset collected from real cameras in China is used as an input to SUMO. IntelliLight is selected as a benchmark in \cite{xu2019targeted}, which introduces a new transfer learning model with a batch learning framework. The same real-world data and a synthetic simulation data which generates traffic with uniform distribution is used on an isolated intersection for experiments. Another DQN-based study for traffic light control with a real dataset is presented in \cite{natafgi2018smart}. Data from a three-way non-homogeneous real intersection in Lebanon is used. The experimental results are compared with the real-world fixed-time controller that is in use at the intersection in terms of queue length and delay.

A different deep RL model in terms of action set and the deep RL algorithm is studied by \textit{Liang et al.}, \cite{liang2019deep}. This work updates the next phase duration in the phase cycle instead of choosing an action from a green phase set. Considering a 4-phase single intersection, phase change duration is defined. The selected phase duration can be added or subtracted from the duration of the next cycle phase. In this model, for a four-way intersection the action set includes 9 discrete actions. The proposed algorithm in this paper considers new DQN techniques, namely double dueling DQN and prioritized experience replay, to improve the performance. In another paper, \textit{Jang et al.} \cite{jang2018agent}, discusses how to integrate a DQN agent with a traffic simulator through the Java-based AnyLogic multipurpose simulator. A different approach for state definitions is proposed by \textit{Liu et al.} \cite{liu2018deep} for examining the impacts of DQN on green-wave patterns in a linear road topology. The experiments are performed only on a Python environment that creates traffic data from a probability distribution without using any traffic simulator. 
Moreover, considering the Dedicated Short-Range Communication (DSRC) technology for vehicle-to-infrastructure (V2I) communication, \textit{Zhang et al.} \cite{zhang2018intelligent} addresses TSC under partial detection of vehicles at an intersection.  Their motivation to study TSC with undetected vehicles comes from the case where not all vehicles use DSRC.

\subsubsection{\bf Multi-agent deep RL}
\label{s:madeepRL}

The first deep RL-based multiple intersection control mechanism is presented in \cite{van2016coordinated}, which defines a new reward function and proposes a coordination tool for multiple traffic lights. The reward definition in this paper considers a combination of specific traffic conditions, namely accidents or jam, emergency stops, and traffic light changes, and the waiting time of all vehicles. The reward function properly penalizes each specific traffic situation. For coordination of multiple intersections to have a high traffic flow rate, this paper uses a transfer planning technique for a smaller set of intersections and links the learning results to a larger set of intersections with the max-plus coordination algorithm. In this work, the benchmark is one of the early coordination-based RL methods proposed in \cite{wiering2000multi}. As expected, the DQN-based coordination method outperforms the earlier standard RL-based method. This paper is expanded to a master thesis \cite{van2016deep} by presenting the results for single agent scenario and different multi-agent scenarios. In \cite{shi2018deep}, similar to \cite{van2016coordinated}, a multi-agent deep RL approach on a 2-by-2 intersection grid model is proposed, in which max-plus and transfer learning are used for reaching the global optimal learning with coordination. This paper differs from \cite{van2016coordinated} mainly by using RNN, in particular LSTM, layers instead of fully connected layers for Q-function approximation. Deep RL approach with RNN structure is shown to result in lower average delay compared to Q-learning and fixed-time control in both low and high traffic demand scenarios. 

\textit{Liu et al.} \cite{liu2017cooperative} introduced a cooperative deep RL model for controlling multiple intersections with multiple agents. The presented algorithm is DQN with a ResNet structure used to form the state space. The reward function penalizes the system based on the driver behavior and waiting time with a BPR function (see Section \ref{s:reward}). Cooperation between agents is assured by sharing the policy with other agents every n-step. Experiments for this study are done using a 2-by-2 intersection model on SUMO. SOTL, Q-learning and DQN are selected as reference points for validating the proposed model.

Multiple traffic intersections can be represented as a network graph in which lane connections between roads form a directed graph. \textit{Nish et al.} \cite{nishi2018traffic} presented a GCN-based neural network structure for the RL agent. GCN is combined with a specific RL algorithm called k-step neural fitted Q-iteration \cite{riedmiller2005neural} that updates the agent in a distributed manner by assigning one agent for each intersection considering the whole network to form the state space. The experiment results show that the GCN-based algorithm decreases the waiting time on all 6 intersections compared to the fixed-time controller and standard CNN-based RL controller. A hierarchical control structure is presented in \cite{zhouerl} for TSC. The lower layer optimizes the local area traffic via intersection control while the top layer optimizes the city-level traffic by tuning the degree of optimization of local areas in the lower layer. In this research, multi-intersection learning is built on threshold values collected from individual intersections. The action set of higher level controller is increasing or decreasing the threshold values that change the sensitivity of each intersection to the neighbor intersections. The learning model in this paper is different from the other deep RL-based intersection controllers such that the model decreases the algorithm complexity in higher level control through a threshold-based mechanism instead of setting the phase cycles or phase duration. 

Cooperative multi-agent deep RL model is investigated in \cite{calvo2018heterogeneous}. Here, an agent with an independent double dueling DQN model supported with prioritized experience replay is assigned to each intersection. In order to improve the coordination performance, a special sampling technique, fingerprint, is used in experience replay. Fingerprint technique estimates Q-functions with neighbor agent's policy via Bayesian inference \cite{foerster2017stabilising}. The proposed model is tested on SUMO with heterogeneous multiple intersections. The results show that the proposed algorithm outperforms the fixed-time controller and the DQN controller without experience replay on several travel demand scenarios.

One of the approaches in multi-agent systems is updating only the critical edges to increase the efficiency. \cite{xu2018network} first identifies important nodes based on multiple criteria with a specific ranking algorithm, CRRank, that creates a trip network using a bidirectional tripartite graph. Based on data and tripartite graph, system ranks the edges based on assigned scores. Once critical intersections are identified, RNN structured DQN agent learns the optimal policy. The model is tested with 20, 50 and 100 intersections on SUMO comparing its results with fixed-time, SOTL, Q-learning and DQN controllers. Recently, a cooperative deep RL method with Q-value transfer is proposed in \cite{ge2019cooperative}. At each intersection, a DQN agent controls the traffic light by receiving Q-values from other agents for learning the optimal policy. The proposed algorithm is supported with extensive experiments on homogeneous and heterogeneous intersections. It is important to have a heterogeneous traffic scenario because all the intersections do not have the same characteristics such as the number of roads and number of lanes. The authors compared their results with two benchmark papers: coordinated Q-learning \cite{van2016coordinated} and distributed Q-learning \cite{araghi2015distributed} approaches.

The work in \cite{casas2017deep} investigates applying the deep deterministic policy gradient (DDPG) algorithm to a city-scale traffic network. The author formulated the TSC problem with DDPG by controlling the phase duration continuously. The model updates phase duration of all network at once by keeping the total phase cycle constant in order to control the synchronization throughout the network. In this work, a specific information called speed score, calculated using the maximum speed on each detector, is considered for forming the state vector. Three traffic scenarios are tested from small to large networks: isolated intersection, 2-by-3 grid intersections, and a Barcelona city-scale map with 43 intersections. The proposed approach achieves higher reward performance than multi-agent Q-learning controller. It is remarkable that actor critic models can be applied large intersection models without any extra multi-agent control technique.
Another DDPG-based deep RL controller for large-scale network is studied by \textit{Genders} in his PhD thesis \cite{genders2018deep}. The system model consists of a parallel architecture with decentralized actors for each intersection and central learners each of which cover a subset of the intersections. The policy determines the duration of the green phase in each intersection. To test the performance of the model, Luxembourg city map is used on SUMO with 196 intersections, which is the largest test environment for RL-based TSC up to now.

A multiple actor-learner architecture considering the A2C algorithm is presented for multiple intersections by \textit{Lin et al.} in \cite{lin2018efficient}. Multiple actors observe different states, and follow different exploration policies in parallel. Since actor-critic approaches are built on advantage functions, authors consider a technique called general advantage estimation function in the learning process \cite{schulman2015high}. The presented experiments are performed on a 3-by-3 intersection grid on SUMO, and the results are compared with the fixed-time controller and actuated controller. 

Independent Q-learning is one of the popular multi-agent RL approaches in literature. \textit{Chu et al.} \cite{chu2019multi} recently expanded this approach to independent A2C for multi-agent TSC. The stability problem is addressed with two methods, fingerprints of neighbor intersections and a spatial discount factor. While the former provides each agent with information regarding the local policies and traffic distributions of neighbor agents, the latter enables each agent to focus on improving the local traffic. 
The network structure in the A2C algorithm is an LSTM-based RNN model. 
Both synthetic traffic network with a 5-by-5 grid and a real network from Monaco City with 30 intersections are used for performance evaluation. 

Systematic learning for large-scale traffic networks is achieved with cooperation in \cite{tan2019cooperative}. A large system is divided into subsets in which each local region is controlled with an RL agent. Global learning is achieved by transferring learning policies to the global agent. For local controllers, authors investigated two deep RL algorithms: value based per-action DQN and actor-critic based Wolpertinger-DDPG \cite{dulac2015deep}. Per-action DQN is similar to the standard DQN algorithm, but differs from DQN by considering state-action pair as an input and generating a single Q value. Wolpertinger-DDPG provides a new policy method based on the k-nearest-neighbor approach using DDPG for large-scale discrete action spaces. 
In experiments, 
three different traffic networks are used, and the
results are compared with a decentralized Q-learning algorithm with linear function approximators, and two rule-based baselines (fixed-time and random-time controllers). 

\textit{Coskun et al.} \cite{cocskun2018deep} expands \cite{mousavi2017traffic}, which use value-based DQN and policy-based standard actor-critic, to multiple intersections using value-based DQN and policy-based A2C. The results of both algorithms following deep learning is consistent with the results of standard RL approaches in terms of average reward per episode, where DQN hits higher average reward than A2C.

%% file: Others.tex
\section{Deep RL for Other ITS Applications}
\label{s:deepRLothers}

\begin{table*}[h!]
\caption{Outline of deep RL approaches for other ITS applications } 
\centering 
\begin{tabular}{l c c c c c} 
\hline\hline 
Case Study & Deep RL method & Target application & Solution & Test  & Result comparison\\ [0.5ex] 
\hline\hline 
Sallab et al. \cite{sallab2017deep}&\makecell{DQN\\DDPG}& Autonomous driving & \makecell{Spatial aggregation, \\ Recurrent temporal \\aggregation} &TORCH game& \makecell{RNN-LSTM\\Kalman-GRNN} \\ [1ex] %
\hline 
Xia et al. \cite{xia2016control}&\makecell{DQN with filtered\\ experience replay} & Autonomous driving & Optimum control & TORCH game &NFQ \cite{riedmiller2005neural}\\ [1ex]
\hline 
Xiong et al. \cite{xiong2016combining}&DDPG& Autonomous driving& Collision avoidance & TORCH game & - \\ [1ex]
\hline 
Sharifzadeh et al. \cite{sharifzadeh2016learning}&DQN& Autonomous driving & Lane changing & Personal simulator& Expert driver\\ [1ex]
\hline 
Hoel et al. \cite{hoel2019combining}&AlphaGo Zero & Autonomous driving & \makecell{Decision planning with \\Monte carlo tree search} & Personal simulator & \makecell{MCTS \\ IDM/MOBIL}\\ [1ex]
\hline 
Hoel et al. \cite{hoel2018automated}&DQN&Autonomous driving& \makecell{Speed change\\Lane change}& Personal simulator & \makecell{CNN-FCNN\\IDM}\\ [1ex]
\hline 
Chae et al. \cite{chae2017autonomous}&DQN& Autonomous breaking &Pedestrian detection& \makecell{PreScan\\ vehicle simulator}&\makecell{Without\\Trauma memory}\\ [1ex]
\hline 
Shi et al. \cite{shi2019driving}&Hierarchical DQN&\makecell{Autonomous driving}& \makecell{Safe gap adjustment\\Lane changing} &Personal simulator& - \\ [1ex]
\hline 
Wang et al. \cite{wang2019lane}&Rule-based DQN&\makecell{Autonomous driving}& Lane changing & Udacity simulator&\makecell{Different policy\\structures}\\ [1ex]
\hline
Ye et al. \cite{ye2019automated}&DDPG&\makecell{Autonomous driving}&\makecell{Lane changing\\Car following}&VISSIM& IDM\\ [1ex]
\hline
Makansis et al. \cite{makantasis2019deep}&DDQN&Autonomous driving&\makecell{Optimum\\highway control}&SUMO&DP\\ [1ex]
\hline
Yu et al. \cite{yu2019distributed}&\makecell{Multi-agent \\Q-learning}&Autonomous driving&\makecell{Coordination graphs}&Personal simulator&\makecell{Expert driver\\Independent\\ Q-learning}\\ [1ex]
\hline
Qian et al. \cite{qian2019deep}&Twin delay DDPG&Autonomous driving&\makecell{Path planning\\}&Personal simulator&\makecell{Expert driver\\DQN}\\ [1ex]
\hline
Zhou et al. \cite{zhou2019development}&DDPG&Autonomus driving&\makecell{Optimum control in\\TSC intersections}& Personal simulator &\makecell{Human driver\\Policy gradient\\DQN}\\ [1ex]
\hline
Osinski et al. \cite{osinski2019simulation}&PPO2 \cite{schulman2017proximal}&Autonomous driving&Optimum control&\makecell{Real world\\CARLA}&\makecell{Continuous \\driving model}\\ [1ex]
\hline
Huang et al. \cite{huang2019autonomous}&DDPG&Autonomous driving&\makecell{Human in the loop \\training}&\makecell{IPG CarMker}&Imitation learning\\ [1ex]
\hline 
Isele et al. \cite{isele2018navigating}&DQN&Autonomous driving&\makecell{Navigating in \\occluded intersections}&SUMO& TTC \cite{minderhoud2001extended}\\ [1ex]
\hline 
Kreidieh et al. \cite{kreidieh2018dissipating}&\makecell{TRPO \cite{schulman2015trust}}&\makecell{Stop-and-go\\ wave dissipation}&Transfer learning& Flow &\makecell{Human driver\\Random policy}\\ [1ex]\hline 
Chalaki et al. \cite{chalaki2019zero} & TRPO & Ramp metering &\makecell{Policy transfer\\ to city scale map\\Adversarial noise injection}& Scaled smart city &\makecell{Human Driver}\\ [1ex]
\hline 
Jang et al. \cite{jang2019simulation} & TRPO & Ramp metering &\makecell{Policy transfer\\ to city scale map}& Scaled smart city& IDM controller \\ [1ex]
\hline 
Belletti et al. \cite{belletti2018expert}& TRPO&Ramp metering &Multi-task conrol&Personal simulator&\makecell{REINFORCE\\PPO \cite{duan2016benchmarking}}\\ [1ex]
\hline 
Chaoui et al. \cite{chaoui2018deep}&DQN&Electric vehicle& \makecell{Energy management with \\multiple batteries}&Personal simulator& - \\ [1ex]
\hline 
Wu et al. \cite{wu2019deep}&DDPG&Hybrid electric bus& \makecell{Adaptive energy management\\ to road conditions}& Personal simulator&\makecell{DQN\\DP} \\ [1ex]
\hline 
Hu et al. \cite{hu2018energy}&DQN&Hybrid electric vehicle&Energy management&\makecell{MATLAB\\ADVISOR \cite{markel2002advisor}}& \makecell{Different training\\ models }\\ [1ex]
\hline
Wu et al. \cite{wu2018differential}&DDPG& Freeway control&Variable speed limit& SUMO&\makecell{Q-learning\\DQN}\\ [1ex]
\hline 
Wu et al. \cite{wu2019ctc}&ES \cite{salimans2017evolution}&Freeway control&\makecell{Ramp meter\\Speed limit\\Lane change}&SUMO&\makecell{No control\\DQN-RM\\TRPO-RM\\DDPG-DVSL}\\ [1ex]
\hline
Pandey et al. \cite{pandey2018multiagent}&\makecell{Sparce cooperative\\Q-learning \cite{kok2006collaborative}}&Toll roads& Dynamic lane management& Personal simulator&\makecell{Density based\\Ratio based\\Random search}\\ [1ex]
\hline 
Pandey et al. \cite{pandey2019deep}&\makecell{Vanilla PG\\Proximal PG \cite{schulman2017proximal}}&Express lane pricing&\makecell{Multi-objective opt.\\Transfer learning}&Personal simulator&Feedback Control\\ [1ex]
\hline
Gunarathna et al. \cite{gunarathna2019dynamic}&\makecell{Multi-agent\\Q-learning}&\makecell{Lane direction \\change}&\makecell{Dynamic \\coordination graphs}&\makecell{New york real\\taxi trips}&\makecell{Different lane\\changing models}\\ [1ex]
\hline
Min et al. \cite{min2019deep}&QR-DQN&Driver assistant&\makecell{Lane keeping\\Lane change\\Acceleration control}&Unity \cite{juliani2018unity}&\makecell{DQN\\DDQN}\\ [1ex]
\hline
Schults et al. \cite{schultz2018deep}&DQN&Traffic simulator&\makecell{Calibrating traffic\\models}& - & -\\ [1ex]
\hline
Bacchiani et al. \cite{bacchiani2019microscopic}&A3C&Traffic simulator&\makecell{Calibrating traffic\\models}& - & -\\ [1ex]
\hline
\end{tabular}
\label{t:otherDRL} 
\end{table*}

Several useful deep RL mechanisms have been introduced for various other applications in ITS. One of the major application areas of AI techniques in ITS is autonomous vehicles, where deep RL occupies a great place in this context. Autonomous controlling is studied from various aspects using deep RL approaches. Ramp metering, lane changing, speed acceleration/deceleration, maneuvering on intersections are some of the various examples studied with deep RL (see Table \ref{t:otherDRL}). 

\subsection{Autonomous Driving}
\label{s:autonomous}

Initial papers presenting deep RL-based control for autonomous vehicles experiment their models on the TORCS game environment \cite{wymann2000torcs}. A control framework proposed by \textit{Sallab et al.} \cite{sallab2017deep} uses two types of deep RL methods, DQN approach with RNNs for discrete action set, and actor-critic based DDPG approach for continuous action domain. The authors experimented the algorithms without using replay memory on TORCS, which led to a faster convergence. \textit{Xia et al.} \cite{xia2016control} studied a control strategy called deep Q-learning with filtered experiences (DQFE) for teaching autonomous vehicle how to drive. The learning performance is shown to outperform the neural fitted Q-learning technique on the TORCS game simulator. 

A continuous control strategy proposed in \cite{xiong2016combining} combines the DDPG algorithm for continuous actions with a safety control strategy. The combination is needed because only relying on past experiences does not provide a safe autonomous vehicle control. \textit{Hoel et al.} \cite{hoel2019combining} introduced an autonomous driving model including planning and learning with Monte Carlo tree search and deep RL. Driving planning is done with Monte Carlo tree search and learning how to drive is done with deep RL agent using the AlphaGO Zero algorithm \cite{silver2017mastering}. 
In that work, the proposed method is compared with a baseline called IDM/MOBIL agent for expert driver behaviours \cite{treiber2000congested}, \cite{kesting2007general}. 

Authors in \cite{ye2019automated} studied car following and lane changing behaviours of autonomous vehicles using DDDP method on VISSIM. Another RL-based autonomous driving policy is described by \textit{Makantasis et al.} \cite{makantasis2019deep} using DDQN with prioritized experience replay in mixed autonomy scenarios. Proposed deep RL-based driving policy is compared with DP-based optimal policy in different traffic densities using SUMO. Deep RL autonomous driving research generally targets individual agents in a mixed autonomy environment or a fully autonomous environment for finding the best driving strategy. However, authors in \cite{yu2019distributed} proposed a multi-agent deep RL approach with dynamic coordination graph. In this study, autonomous vehicles in coordination learn how to behave in a highway scenario. Two distinct coordination graph models, identity-based dynamic coordination and position-based dynamic coordination, are studied in that work. \textit{Qian et al.} \cite{qian2019deep} described autonomous driving from a different perspective using twin delayed DDPG \cite{fujimoto2018addressing}. They proposed a two-level strategy to fill the gap between decision making and future planning of autonomous vehicle. Autonomous driving in a signalized traffic intersection using DDPG method is proposed by \textit{Zhou et al.} \cite{zhou2019development}. In a recent autonomous driving study \cite{osinski2019simulation}, RL methods are analyzed on the traffic simulator CARLA \cite{dosovitskiy2017carla} using RGB image inputs collected from a camera. A different training and test strategy is experimented by authors in \cite{huang2019autonomous} for DDPG-based autonomous driving using a human-in-the-loop dynamical simulator called IPG CarMaker. While a human driver controls the vehicle on this software, the DDPG agent learns how to drive in two distinct scenarios, forward driving and stopping.

In transportation research, controlling stop-and-go waves with autonomous vehicles is a new approach for which a deep RL-based solution is suggested in \cite{kreidieh2018dissipating}. The authors implemented multiple autonomous vehicles controlled by individual deep RL agents to increase the flow of traffic. \textit{Isele et al.} \cite{isele2018navigating} using the DQN approach studied a special case for self-driving vehicles, maneuvering in intersections when driver have partial knowledge about the intersection. In this paper, three action selection modes are tested. First action mode is stop or go, the second mode is having sequential actions, accelerate, decelerate or keep constant velocity, and the last action mode is the combination of first two action modes, wait, move slowly or go. All three action modes are tested on 5 different cases. 

The authors in \cite{hoel2018automated} proposed a speed and lane changing framework for autonomous truck-trailer with surrounded vehicles using double DQN. This work considers several traffic situations including a highway traffic and a two-way traffic scenario called overtaking in order to generalize the proposed algorithm. Using an inverse deep RL approach, \textit{Sharifzadeh et al.} \cite{sharifzadeh2016learning} presented a driving model for collision-free lane changing on a self-programmed traffic simulator with continuous trajectories. The investigated model includes two separate agents. One agent controls only lane changing without speed adjustment, and the other agent controls lane changing actions with acceleration. Another lane changing application for autonomous vehicles is presented in \cite{shi2019driving} considering DQN with quadratic Q-function approximator. A hierarchical control technique is implemented as a lane changing module in discrete domain, and a gap adjusting module in continuous domain with separate deep RL agents. Similar to the other papers, authors in \cite{wang2019lane} proposed a rule-based DQN approach for the lane changing problem for autonomous vehicles. 

Most of the learning based control models in ITS test the proposed work on simulators such as autonomous vehicle control, traffic signal control, traffic flow control. The first learned policy transfer from simulator to real world experiments is studied by \textit{Chalaki et al.} \cite{chalaki2019zero}. Experiment platform for this research is a scaled city map from University of Delaware, in which behaviors of multiple autonomous vehicles in a roundabout is observed with deep RL control techniques. In order to transfer policies efficiently, adversarial noise is injected into the state and action spaces. The initial results of the same work for single agents with Gaussian noise is studied in \cite{jang2019simulation}.

\subsection{Energy Management}
\label{s:energy}

Energy management systems are a crucial part of future transportation. There are different resource allocation schemes for electric vehicles. Power consumption varies in different units of vehicle that highly effects the performance of batteries. \textit{Chaoui et al.} proposed a deep RL-based energy management solution to increase the life cycle of parallel batteries \cite{chaoui2018deep}. Authors in \cite{hu2018energy} suggest an optimization model for energy consumption in hybrid vehicles using the DQN formulation. Proposed adaptive learning model provides a better fuel consumption through deep RL-based energy management scheme. \textit{Wu et al.} \cite{wu2019deep} proposed an energy management solution for hybrid electric buses using an actor-critic based DDPG algorithm. Considering two parameters, number of passengers and traffic information, deep RL agent can optimize the energy consumption with continuous controlling. 

\subsection{Road Control}
\label{s:road}

Road controllers are an essential part of traffic control in ITS. There are several works which use deep RL methods for speed limit control, toll road pricing, ramp metering, etc.
Dynamic speed limit control among lanes is a challenging task in transportation. \textit{We et al.} \cite{wu2018differential} studies a dynamic solution method with actor-critic continuous control scheme for variable speed limits control, that increases the flow rate and decreases the emission rate. Deep RL-based lane pricing model for toll roads is proposed in \cite{pandey2018multiagent} to maximize the total revenue with multiple entrance and exits.  
Another dynamic lane pricing model for express lanes is proposed in \cite{pandey2019deep}, where authors used multi-objective RL and multi-class cell transmission models to enhance the performance of deep RL agent.
Highway connections from side roads are controlled with signalized ramp meters. In order to increase the efficiency of the main road flow, a new multi-agent deep RL technique is proposed in \cite{belletti2018expert} for traffic models based on discretized partial differential equations. The control model is tested on a simulated highway scenario with multiple ramp meters. 
\textit{Wu et al.} \cite{wu2019ctc} proposed a freeway control model using deep RL with various agents for different parts of freeway. Authors' proposal is to use an inflow ramp meter control agent, a dynamic lane speed limit control agent, and a dynamic lane change controller agent in coordination. Traditional roads have fixed number of lanes for incoming and outgoing directions. Lane direction change is studied for improving the traffic flow with multi-agent deep RL and dynamic graph configuration in \cite{gunarathna2019dynamic}.
Autonomous braking system via DQN is proposed in \cite{chae2017autonomous}, which provides traffic safety in cases where immediate action is required.

\subsection{Various ITS Applications}
\label{s:various}

Recently a new tool for optimizing traffic simulators is proposed by \textit{Schultz at al.} \cite{schultz2018deep}. The input, (traffic characteristics) and output (traffic congestion) of traffic simulators are correlated with an adaptive learning technique using DQN. Another computational interface, named Flow, enables easy integration of the deep RL library RLlib \cite{liang2018rllib} with SUMO and Aimsun for various control problems in ITS \cite{wu2017flow}. \textit{Flow} users can create a custom network via Python to test complex control problems such as ramp meter control, adaptive traffic signalization and flow control with autonomous vehicles. Authors in \cite{bacchiani2019microscopic} introduces a traffic simulator which provides a new environment with cooperative multi-agent learning approach for analyzing the behaviours of autonomous vehicles. It is capable of testing various traffic scenarios. 
\textit{Min et al.} \cite{min2019deep} proposed a driver assistant system using quantile regression DQN for various controls such as lane keeping, lane changing, and acceleration control.

\section{Challenges and Open Research questions}
\label{s:challenges}

Despite the significant interest and effort, and the promising results so far in deep RL-based ITS solutions, there are still many major challenges to address before the proposed research can yield real-world products. In this section, we will discuss the major challenges and open research questions of deep RL for ITS.

All the research outcomes for RL-based ITS controls are experimented on simulators due to life threatening consequences of real-world applications. Recently, authors in \cite{chalaki2019zero} presented a policy transfer application from simulation to a city-scale test environment for autonomous driving, but still this line of research is in its infancy. There is a huge gap between real-world deployment and simulator-based applications using learning algorithms. For TSC and other controlling applications in ITS, a real-world deployment is needed in order to prove the applicability of deep RL-based automated control.

Specifically for TSC, simulation-based applications have two approaches in literature, first, simulating an artificial road network with artificial data, second, simulating a road network based on a real dataset. While the second one is close to a realistic test, it only considers the traffic demand in various times of the day without realistic challenges. Another point that researchers need to consider for TSC is increasing the realism of simulation environments, such as including the human intervention scenarios. In order to decrease human intervention in TSC, the control system should be adaptable to unstable traffic situations in the worst case scenarios. To do that, instead of standard traffic models, urban networks with some predictable extreme scenarios should be studied in order to see the consequences of deep RL implementations. We expect that implementing pedestrians and public transportation to the simulation environments will have a high impact on learning performance.  

There are so many proposed deep RL models in the literature for controlling traffic lights. While standard RL models have comparisons between each other for validating their proposals, deep RL models on TSC do not have satisfactory comparisons with existing works. For multiple intersections, researchers mostly selected DQN, standard RL, and fixed-time controllers as benchmark. However, they should be especially compared with other multi-agent approaches in the literature, such as distributed control, coordinated control, etc. Another challenge with results is that very few papers compare their performance with actuated controller, which is the most popular real-world TSC implementation.

State definition is a crucial point in deep RL applications. Thus, researchers pay attention to different state forms with different hardware systems such as cameras, loop detectors, and sensors, but still there is no clear agreement on the form of state in deep RL-based TSC applications. State definition highly depends on static devices, hence all of them should always collect data properly. A new research direction could be studying partially observable and noisy state definitions in which some of the devices do not work properly. When RL-based adaptive traffic signals are implemented on intersections, the system must be protected and stable (i.e., robust and resilient) against such kind of failures.

Regarding autonomous vehicles, researchers have been proposing solutions to very specific subsystems without considering the interaction between such subsystems. For more realistic solutions, a unified management and adaptive control strategy is required for several components. For example, an impactful deep RL application should control lane changing, breaking, flow arranging, and energy management components all together. Implementing different learning algorithms for different autonomous vehicle subsystems may cause interoperability issues.

\section{Conclusion}
Considering the increasing world population and urbanization, researchers have been conducting research on ITS applications using learning-based AI techniques.
Dynamic nature of traffic systems does not allow a clear easy control mechanism for all ITS applications. 
Controlling transportation systems through reinforcement learning (RL) approaches is gaining popularity in both industry and academia. There are various research outcomes in recent years for solving automated control problems in ITS, such as traffic lights, autonomous driving, autonomous break, and energy management of vehicles. The most popular deep RL application in ITS 
is adaptive traffic signal control (TSC) at intersections.
We presented a comprehensive review for the deep RL applications in ITS.
Key concepts of RL and deep RL, and the settings in which they are applied to TSC were discussed to provide a smooth introduction to the literature.
Characteristic details of existing works in several categories were compared in separate tables in order to enable clear comparison.
Finally, we also discussed the open research directions and the gap between the existing works and the real-world usage.   
This survey showed that there are different single agent and multi-agent RL solutions for TSC that outperform the standard control methods in simulation environments. However, existing works have still not been tested in real-world environments except for an autonomous vehicle application for a specific scenario. 